\RequirePackage{iftex}
\ifPDFTeX
  \pdfoutput=1
\else
  \ifXeTeX
    \PassOptionsToPackage{xetex}{hyperref}
  \fi
\fi
\documentclass[11pt]{article}

\PassOptionsToPackage{hyperfootnotes=false}{hyperref}
\usepackage[preprint]{acl}

\usepackage{times}
\usepackage{latexsym}
\usepackage[T1]{fontenc}
\usepackage[utf8]{inputenc}
\usepackage{microtype}
\usepackage{inconsolata}

\usepackage{enumitem}
\usepackage{amsmath,amssymb}
\usepackage{booktabs}
\usepackage{graphicx}
\usepackage{float}
\usepackage{stfloats}
\usepackage{array}
\usepackage{makecell}
\usepackage{ragged2e}
\usepackage{siunitx}
\usepackage{tabularx}
\usepackage{algorithm}
\usepackage[noend]{algpseudocode}
\algrenewcommand\algorithmicrequire{\textbf{Input:}}
\algrenewcommand\algorithmicensure{\textbf{Output:}}
\algrenewcommand\algorithmiccomment[1]{\hfill{\footnotesize\itshape #1}}
\usepackage[capitalize]{cleveref}
\usepackage{CJKutf8}
\usepackage{placeins}
\usepackage{url}

\graphicspath{{figures/}}

\newif\ifshowidentity
\newif\ifpreprintmode
\newif\iffrontpanel
\newif\ifpanel
\makeatletter
\ifacl@anonymize\showidentityfalse\else\showidentitytrue\fi
\ifacl@finalcopy\ifacl@pagenumbers\preprintmodetrue\fi\fi
\makeatother
\iffrontpanel\ifpreprintmode\paneltrue\fi\fi

\usepackage{fancyhdr}
\fancypagestyle{preprintfirst}{%
  \fancyhf{}%
  \fancyfoot[L]{\small Accepted to Findings of EACL 2027.}%
  \fancyfoot[C]{\small\thepage}%
}
\ifpanel
\usepackage{environ}
\usepackage{tcolorbox}
\definecolor{frontpanelgrey}{gray}{0.955}
\makeatletter
\RenewEnviron{abstract}{%
  \twocolumn[{%
    \begin{tcolorbox}[colback=frontpanelgrey,colframe=frontpanelgrey,
        boxrule=0pt,arc=12pt,left=16pt,right=16pt,top=14pt,bottom=14pt]
      {\raggedright\fontfamily{qpl}\selectfont\LARGE\bfseries \@title\par}
      \vspace{12pt}
      {\raggedright\fontfamily{qpl}\selectfont\bfseries \preprintauthorline\par}
      \vspace{3pt}
      {\raggedright\fontfamily{qpl}\selectfont \preprintaffiliation\par}
      \vspace{8pt}
      \noindent\BODY\par
      \vspace{8pt}
      \noindent
      \begin{minipage}[c]{0.82\linewidth}
        \raggedright
        \textbf{Correspondence:} \href{mailto:\preprintemail}{\texttt{\preprintemail}}\\[2pt]
        \textbf{Code:} \preprintcodelink\\[2pt]
        \textbf{Dictionaries:} \preprintdictlink
      \end{minipage}%
      \hfill
      \begin{minipage}[c]{0.14\linewidth}\raggedleft\preprintlogos\end{minipage}
    \end{tcolorbox}
    \vskip 0.8\baselineskip
  }]%
  \thispagestyle{preprintfirst}%
}
\makeatother
\fi

\ifshowidentity
  \AtBeginDocument{\hypersetup{
    pdftitle={Sparse Readout Prism: Explaining Logit-Lens Scores in Features Instead of Tokens},
    pdfauthor={Matteo He, William F. Shen, Xinchi Qiu, Nicholas D. Lane}}}
\fi

\newcommand{\blfootnote}[1]{%
  \begingroup
    \renewcommand{\thefootnote}{}%
    \footnotetext{#1}%
  \endgroup
}

\setlist[itemize]{leftmargin=*,nosep}
\setlist[enumerate]{leftmargin=*,nosep}
\setlist[description]{nosep}
\renewcommand{\topfraction}{0.94}
\renewcommand{\bottomfraction}{0.85}
\renewcommand{\dbltopfraction}{0.94}
\renewcommand{\textfraction}{0.06}
\renewcommand{\floatpagefraction}{0.82}
\renewcommand{\dblfloatpagefraction}{0.82}

\newcolumntype{Y}{>{\RaggedRight\arraybackslash}X}
\newcolumntype{Z}{>{\Centering\arraybackslash}X}

\makeatletter
\newcommand{\AppendixFloatSetup}{%
  \setlength{\@fptop}{0pt}%
  \setlength{\@fpsep}{10pt plus 1fil}%
  \setlength{\@fpbot}{0pt plus 1fil}%
  \setlength{\@dblfptop}{0pt}%
  \setlength{\@dblfpsep}{10pt plus 1fil}%
  \setlength{\@dblfpbot}{0pt plus 1fil}%
  \setlength{\textfloatsep}{0.52\baselineskip plus 2pt minus 2pt}%
  \setlength{\dbltextfloatsep}{0.58\baselineskip plus 2pt minus 2pt}%
  \setlength{\floatsep}{0.45\baselineskip plus 2pt minus 2pt}%
  \setlength{\intextsep}{0.50\baselineskip plus 2pt minus 2pt}%
  \setlength{\abovecaptionskip}{0.35\baselineskip plus 1pt minus 1pt}%
  \setlength{\belowcaptionskip}{0.08\baselineskip plus 1pt minus 1pt}%
  \setcounter{topnumber}{5}%
  \setcounter{bottomnumber}{3}%
  \setcounter{totalnumber}{8}%
  \setcounter{dbltopnumber}{3}%
  \setcounter{dblbotnumber}{2}%
  \renewcommand{\topfraction}{0.97}%
  \renewcommand{\bottomfraction}{0.90}%
  \renewcommand{\dbltopfraction}{0.97}%
  \renewcommand{\textfraction}{0.03}%
  \renewcommand{\floatpagefraction}{0.80}%
  \renewcommand{\dblfloatpagefraction}{0.80}%
}
\makeatother

\newcommand{\AppendixCompactGraphic}[2][]{%
  \includegraphics[width=\linewidth,height=0.32\textheight,keepaspectratio,#1]{#2}%
}

\newcommand{\AppendixWideGraphic}[2][]{%
  \includegraphics[width=\textwidth,height=0.44\textheight,keepaspectratio,#1]{#2}%
}

\newcommand{\WU}{W_U}
\newcommand{\R}{\mathbb{R}}
\newcommand{\dmodel}{d_{\mathrm{model}}}
\newcommand{\rstate}{\tilde h_\ell}   % decoded state at layer \ell
\newcommand{\rstateL}{\tilde h_L}     % decoded state at the final layer
\newcommand{\sexact}{s_{\mathrm{exact}}}
\newcommand{\srecon}{s_{\mathrm{recon}}}

\newcommand{\mexact}{m_{\mathrm{exact}}}
\newcommand{\mrecon}{m_{\mathrm{recon}}}

\ifpanel
\title{\raisebox{-0.6em}{\includegraphics[height=1.95em]{figures/schematic_tikz/prism_icon.pdf}}%
       \hspace{0.55em}%
       \begin{tabular}[c]{@{}l@{}}
         Sparse Readout Prism: Explaining Logit-Lens \\
         Scores in Features Instead of Tokens
       \end{tabular}}
\else
\title{Sparse Readout Prism: Explaining Logit-Lens Scores\\in Features Instead of Tokens}
\fi

\newcommand{\preprintauthorline}{Matteo He, William F.\ Shen,
  Xinchi Qiu, Nicholas D.\ Lane}
\newcommand{\preprintaffiliation}{Department of Computer Science and Technology,
  University of Cambridge, Cambridge, United Kingdom}
\newcommand{\preprintemail}{mh2274@cantab.ac.uk}
\newcommand{\preprintcodelink}{\href{https://github.com/hematteo/sparse-readout-prism}{\texttt{github.com/hematteo/sparse-readout-prism}}}
\newcommand{\preprintdictlink}{\href{https://huggingface.co/hematteo/sparse-readout-prism}{\texttt{huggingface.co/hematteo/sparse-readout-prism}}}
\newcommand{\preprintlogos}{\includegraphics[height=30pt]{logos/cambridge_crest}}

\author{
  Matteo He\textsuperscript{\dag} \qquad William F. Shen \qquad
  Xinchi Qiu \qquad Nicholas D. Lane \\[10pt]
  Department of Computer Science and Technology, University of Cambridge \\[1pt]
  Cambridge, United Kingdom
}

\begin{document}
% With the front panel on (\ifpanel, style/preamble.tex) it replaces
% \maketitle and carries the correspondence line. Otherwise every build uses
% the standard ACL title, the final and preprint builds key the author marks,
% and the preprint build adds its first-page Findings footer.
\ifpanel\else
\maketitle
\fi
\ifpreprintmode\ifpanel\else\thispagestyle{preprintfirst}\fi\fi
\ifshowidentity\ifpanel\else
% Two key lines with aligned text: the dagger hangs left of the first line,
% and the artifact links share the second line.
\blfootnote{\llap{\textsuperscript{\dag}}Corresponding author: \texttt{mh2274@cantab.ac.uk}.}
\blfootnote{\href{https://github.com/hematteo/sparse-readout-prism}{Code on GitHub};
  \href{https://huggingface.co/hematteo/sparse-readout-prism}{SAEs on Hugging Face}.}
\fi\fi

\begin{abstract}
Lens methods interpret intermediate hidden states by transporting them to
the final layer and decoding them through the unembedding matrix into token
scores.
Fitted lenses learn their transport map from a training corpus.
We find that changing only the corpus can change token rankings for the
same hidden state; we call this corpus conditionality.
To test whether two differently fitted lenses share unembedding features,
we introduce Sparse Readout Prism (SRP), which learns these features by
reconstructing each unembedding row as a sparse combination of them.
SRP measures each feature's contribution to a lens score without changing
the lens's transport.
Across six models, SRP reconstructs logit differences more accurately than
geometric baselines.
On Ministral-3-8B, English and Chinese Jacobian lenses disagree on the
top token for all prompts at one layer in both comparisons, versus at most
10\% for same-language refits at that layer.
On Qwen3.5-9B, disagreement reaches 87\% between English and German
Jacobian lenses and 94\% between English and Chinese ridge lenses.
At one layer, SRP finds the same dominant feature in the two lenses'
top-token scores for 81\% of prompts with English and German Jacobian
lenses on Qwen and up to 78\% with English and Chinese Jacobian lenses
on Ministral.
% The preprint's code and dictionary links moved to the front panel's
% metadata strip (style/preamble.tex); the review build never showed them.
\end{abstract}

\section{Introduction}

\begin{figure}[t]
    \centering
    \includegraphics[width=\columnwidth]{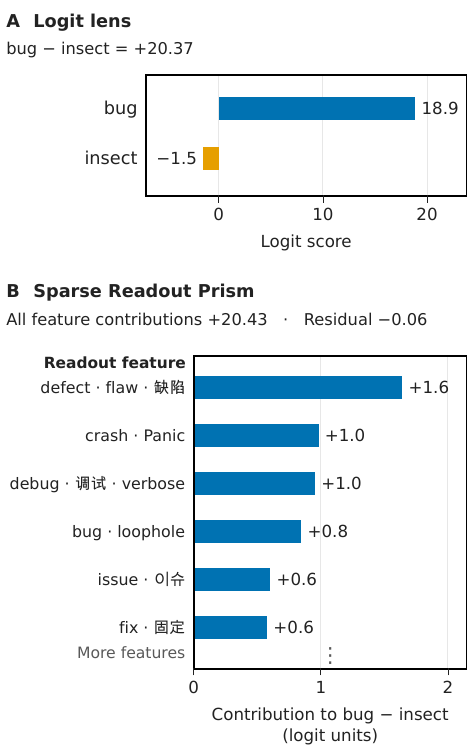}
    \caption{\textbf{Where a lens reading comes from.}
    \textbf{(A)} Logit lens scores for \texttt{bug} and \texttt{insect}
    in Qwen3.5-2B on ``The programmer reproduced the crash and filed a''.
    \textbf{(B)} SRP decomposes their difference into signed feature
    contributions in logit units. The leading features concern software
    defects. Each label shows selected tokens from that feature's top
    unembedding rows. The plot omits small contributions and features
    associated with casing, spacing, or punctuation. The reported sum
    includes all feature contributions. Adding the residual recovers the
    score difference in panel A.}
    \label{fig:main-lens-prism-comparison}
\end{figure}

Lens methods provide a view of intermediate transformer computation
through predictions over the model's vocabulary
\citep{nostalgebraist2020,belrose2023tuned,pal2023future,ghandeharioun2024patchscopes}.
These readings have been used to trace how predictions develop across
layers, inspect factual information in hidden states, and investigate
intermediate reasoning
\citep{belrose2023tuned,ghandeharioun2024patchscopes,gurnee2026workspace}.

A lens prepares an intermediate state for decoding, then scores tokens
with the unembedding matrix \(\WU\), also called the LM head. We refer to
this matrix as the readout. Each row of \(\WU\) assigns a score to one
vocabulary token. The plain logit lens applies the model's output
normalization to the intermediate state and scores it with the original
head. Fitted lenses, including the tuned lens
\citep{belrose2023tuned} and the Jacobian lens
\citep{gurnee2026workspace}, use a corpus to estimate a transport map
that transforms the state before decoding.

Figure~\ref{fig:main-lens-prism-comparison} illustrates the ambiguity
of a token-level explanation. The logit lens scores \texttt{bug} above
\texttt{insect} on a prompt about software, but every sense of \texttt{bug}
shares one row of the LM head, so the score does not separate contributions
associated with its different senses. Near-synonyms and translation
equivalents occupy different readout rows, so a token-level account can
miss what their predictions have in common. Token identity is too coarse to
distinguish senses of the same token and too fine to identify structure
shared across tokens.

The corpus used to fit a lens can also shape which tokens it predicts, a
dependence we call corpus conditionality. This matters when researchers use
token predictions to infer whether intermediate computation proceeds in English
\citep{wendler2024llamas,schut2025english,gurnee2026workspace}.
We find that lenses fitted on English and Chinese text can predict tokens in
different languages from identical hidden states (\S\ref{sec:beyond-token}).
The language of those tokens alone therefore does not establish the language
of the intermediate computation.

Sparse Readout Prism (SRP) explains lens scores through shared readout
features. We learn these features by fitting a sparse autoencoder to the
unembedding rows. Each row can combine several features, and each feature
can contribute to multiple rows. Given the state a lens passes to the LM
head, SRP measures each feature's signed contribution to a token logit or
logit difference and reports the reconstruction residual explicitly
(\S\ref{sec:method-local-score}).

We make two contributions.
\begin{enumerate}[leftmargin=*, itemsep=2pt, topsep=3pt]
\item SRP (\S\ref{sec:method}) explains lens scores through
contributions from shared readout features. Across six models, it
increases reconstruction coverage by 8.9 to 17.3 percentage points over
the strongest of six baselines built on row geometry. A small set of SRP
features also distinguishes word senses in held-out contexts
(\S\ref{sec:results}).
\item We demonstrate corpus conditionality in fitted lenses on two models
and use SRP to identify shared structure and differences across their
readings (\S\ref{sec:beyond-token}).
\end{enumerate}

\section{Related Work}

\paragraph{Lens methods.}
Lens methods extend direct vocabulary projections
\citep{nostalgebraist2020} with fitted decoding
\citep{belrose2023tuned}, prediction of future tokens
\citep{pal2023future}, generated explanations
\citep{ghandeharioun2024patchscopes}, and estimates of verbalizable
content \citep{gurnee2026workspace}. SRP complements these methods by
treating shared readout features as units of analysis. These features
are learned independently of the lens and provide a common basis for
comparing readings that predict different tokens.

\paragraph{Sparse dictionaries and parameter interpretation.}
Activation sparse autoencoders (SAEs) learn interpretable features from
hidden states collected over a corpus
\citep{bricken2023monosemanticity,cunningham2023sparse,gao2024scaling,templeton2024scaling},
while SRP learns features directly from the rows of the unembedding matrix.
Crosscoders extend activation dictionary learning to shared features
across layers and models \citep{lindsey2024crosscoders}. Static word
embeddings, like unembedding rows, assign fixed vectors to vocabulary
items. SPINE learns sparse features from GloVe and word2vec, and sparse
coding recovers word-sense components
\citep{subramanian2018spine,arora2018polysemy}. Parameter decomposition
likewise learns mechanistic components of model weights
\citep{braun2025parameter}, while weight-based SAE analyses characterize
features already learned from activations \citep{liu2026weightbasedsae}.
Closely related work fits a shared sparse dictionary to unembedding rows
across pretraining checkpoints
\citep{he2026learningreadoutunembedding}. That work studies readout
development during training, while SRP learns its dictionary directly
from a fixed unembedding to explain and compare individual lens scores
through contributions from shared readout features.

\section{Sparse Readout Prism}
\label{sec:method}

\begin{figure}[t]
    \centering
    \includegraphics[width=\columnwidth]{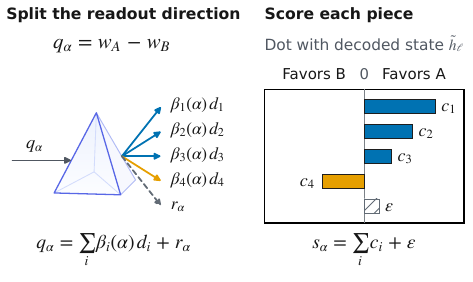}
    \vspace{-0.8\baselineskip}
    \caption{\textbf{SRP decomposes lens scores into signed contributions
    from shared readout features.} Here, the score is the logit difference
    between tokens \(A\) and \(B\). The left panel splits the corresponding
    readout direction into weighted feature directions and a residual.
    The right panel shows how much each component contributes for the
    decoded state. Blue bars favor \(A\), orange bars favor \(B\), and the
    hatched bar gives the residual contribution. All contributions are in
    logit units and sum exactly to the logit difference
    (\S\ref{sec:method-local-score}).}
    \label{fig:method-schematic}
\end{figure}

SRP explains a token logit or logit difference through contributions
from shared readout features (\Cref{fig:method-schematic}). It first
learns a sparse dictionary from the rows of the LM head. For a decoded
state, it then computes how much each feature supports or opposes the
chosen score. These contributions, together with a shared offset and a
reconstruction residual, sum exactly to the score. The dictionary remains
fixed across contexts, layers, and lenses, providing common features for
comparing their readings.

\paragraph{Notation.}
Throughout, \(h_\ell\in\R^{\dmodel}\) denotes the hidden state at layer
\(\ell\) before a lens decodes it, and \(\dmodel\) is the
dimension of the model's hidden states. The map
\(T_\ell\) combines the operations that transform the hidden state
before it reaches the LM head, including the model's output normalization
and the lens's transport in the order used by that lens.
We write \(\rstate=T_\ell(h_\ell)\in\R^{\dmodel}\) for the
decoded state passed to the LM head. The unembedding matrix
\(\WU\in\R^{|\mathcal{V}|\times\dmodel}\) over the vocabulary
\(\mathcal{V}\) then produces the logits \(\WU\rstate\). Row \(v\),
written as the column vector \(w_v:=\WU[v,:]^\top\), gives token \(v\)
the linear score
\[
    \operatorname{logit}_v(\rstate)=\rstate^\top w_v.
\]
SRP factorizes \(\WU\) and decomposes scores for whatever state the lens
passes to the head, without assuming that intermediate states already
align with the readout.

\subsection{Factorizing the LM Head into Sparse Readout Features}
\label{sec:method-factorization}

SRP fits a sparse autoencoder (SAE) to the rows of
\(\WU\).
The learned dictionary expresses each token's unembedding row as
\(w_v = \widehat{w}_v + r_v\) with
\(\widehat{w}_v = \mu + \textstyle\sum_{i=1}^{D} z_{v,i} d_i\),
where \(\widehat{w}_v\) is the fitted row, \(r_v\) is the residual,
\(\mu\) is the mean of the rows used to fit the SAE, \(D\) is the dictionary width,
\(d_i\in\R^{\dmodel}\) is a readout feature direction, and
\(z_{v,i}\) is the sparse coefficient of row \(v\) on feature \(i\).
Other sparse dictionary learners can supply the same form of row factorization.

The fitting stage in \Cref{alg:srp} forms the SAE input \(x_v\) by subtracting
\(\mu\) from each row and dividing by the centered row norm \(a_v\), floored at
\(10^{-8}\) in the implementation.
We use TopK sparse autoencoders~\citep{makhzani2013ksparse,gao2024scaling}.
The encoder applies ReLU and keeps the \(k\) largest activations in each
row, so each row uses at most \(k\) features. The main reconstruction
experiments use \(k=256\) at 32\(\times\) expansion
(Appendix~\ref{app:readout-factorizer-qwen-sweeps}).

We train the SAE to reconstruct the unembedding rows while keeping the top
token's logit close to its original value.
For the reconstruction term, we use the mean squared error between the
sampled rows \(x_v\) and their reconstructions \(\widehat{x}_v\)
\[
    \mathcal L_{\mathrm{row}} = \frac{1}{\dmodel}\,
    \mathbb E_v\big[\lVert x_v-\widehat{x}_v\rVert_2^2\big].
\]
But equal row errors can produce different logit errors, depending on how
the residuals project onto the LM head input. So we add a logit term for
the token \(v_*\) that receives the highest logit under the original head at
each final-layer input \(\rstateL\) sampled from English text.
We restore the original row scale using \(\widehat{w}_v=\mu+a_v\widehat{x}_v\)
and penalize the squared relative error in that token's logit
\[
    \mathcal L_{\mathrm{logit}} = \mathbb E_{\rstateL}\!\left[
    \left(\frac{\rstateL^\top r_{v_*}}
    {\max\{\lvert\rstateL^\top w_{v_*}\rvert,10^{-5}\}}\right)^2\right].
\]
The \(10^{-5}\) floor keeps the denominator away from zero.
The full objective combines the row loss with the weighted logit loss and
a small squared \(\ell_2\) penalty on the sparse codes.
Appendix~\ref{app:model-suite-selection} gives the training recipe and
dictionary selection criteria.

\begin{algorithm}[t]
\caption{Sparse Readout Prism}
\label{alg:srp}
\small
\begin{algorithmic}[1]
\Require rows \(\{w_v\}\) of \(\WU\), dictionary learner \(\mathcal{S}\), width \(D\)
\Function{Fit}{$\{w_v\}$}\Comment{once per model}
\State \(\mu\gets\) mean unembedding row
\State \(a_v\gets\lVert w_v-\mu\rVert,\;\; x_v\gets(w_v-\mu)/a_v\)
\State \(\big(\{z^{\mathrm{pre}}_{v,i}\},\{d_i\}_{i=1}^{D}\big)\gets\mathcal{S}(\{x_v\})\)\Comment{here TopK}
\State \(z_{v,i}\gets a_v\,z^{\mathrm{pre}}_{v,i}\)
\State \(r_v\gets w_v-\mu-\sum_i z_{v,i}\,d_i\)
\State \Return \(\mu,\{d_i\},\{z_{v,i}\},\{r_v\}\)
\EndFunction
\Statex
\Function{Decompose}{$\alpha,\rstate$}\Comment{once per score}
\State \(\beta_i\gets\sum_v\alpha_v z_{v,i}\) for \(i=1,\dots,D\)
\State \(c_i\gets\beta_i\,\rstate^\top d_i\)\Comment{feature contributions}
\State \(o\gets\big(\sum_v\alpha_v\big)\,\rstate^\top\mu\)\Comment{offset, 0 for contrasts}
\State \(\epsilon\gets\rstate^\top\sum_v\alpha_v r_v\)\Comment{residual}
\State \Return \(o,\{c_i\},\epsilon\)\Comment{\(s_\alpha(\rstate)=o+\sum_i c_i+\epsilon\)}
\EndFunction
\end{algorithmic}
\end{algorithm}

\subsection{Decomposing Selected Readout Scores}
\label{sec:method-local-score}

Token logits and margins are the simplest scores we evaluate.
For two tokens \(A,B\in\mathcal{V}\), the margin is the signed logit
difference \(m_{A,B}(\rstate)=\rstate^\top(w_A-w_B)\).

We select the score using a coefficient vector
\(\alpha\in\R^{|\mathcal{V}|}\) over unembedding rows. This defines the
selected readout direction
\(q_\alpha = \textstyle\sum_{v\in\mathcal{V}} \alpha_v w_v\) and the
corresponding score \(s_\alpha(\rstate)=\rstate^\top q_\alpha\).
Setting \(\alpha_A=1\) and all other coefficients to zero recovers the
token logit \(\operatorname{logit}_A(\rstate)=\rstate^\top w_A\) for a
vocabulary token \(A\in\mathcal{V}\). For the margin with \texttt{bug} as
\(A\) and \texttt{insect} as \(B\), set \(\alpha_B=-1\) as well.

The coefficient of readout feature \(i\) in the selected direction is
\(\beta_i(\alpha)=\textstyle\sum_{v\in\mathcal{V}} \alpha_v z_{v,i}\).
With the readout feature directions \(d_i\) fixed, the projection
\(p_i(\rstate)=\rstate^\top d_i\) measures how strongly the decoded
state aligns with feature \(i\). Their product
\(c_i(\rstate,\alpha)=\beta_i(\alpha)\,p_i(\rstate)\) is the signed
feature contribution to the score.

\begin{samepage}
For a selected direction \(q_\alpha\), define its reconstruction residual
\[
    r_\alpha=\textstyle\sum_{v\in\mathcal{V}}\alpha_v r_v.
\]
\end{samepage}
Substituting the row factorizations expresses the score as a shared
offset, the sum of feature contributions, and a residual
\[
\begin{aligned}
    s_\alpha(\rstate)
    &=
    \big(\textstyle\sum_v \alpha_v\big) \rstate^\top\mu \\
    &\quad + \textstyle\sum_{i=1}^{D}
        \big(\textstyle\sum_v \alpha_v z_{v,i}\big) \rstate^\top d_i \\
    &\quad + \rstate^\top r_\alpha
\end{aligned}
\]
We write \(\srecon\) for the reconstructed score, which sums the
shared offset and signed contributions above. The exact score is
\(\sexact=s_\alpha(\rstate)\), and the residual term is
\(\epsilon=\sexact-\srecon
=\rstate^\top r_\alpha\).
The residual need not be orthogonal to the dictionary directions, and
its signed score term records what the reconstruction leaves unexplained.
Every display shows the residual term \(\epsilon\) beside the
contributions.

A selected score whose coefficients sum to zero is a contrast, and
every margin is a contrast. For such scores, the offset term
\(\left(\sum_v\alpha_v\right) \rstate^\top\mu\) vanishes.

Appendix~\ref{app:score-decomposition} gives \(q_\alpha\) and
\(\beta_i(\alpha)\) for each score family we use.

\subsection{Comparing Decompositions and Assessing Fidelity}
\label{sec:method-comparing-accounts}
\label{sec:method-reading-local-accounts}

Within a model, SRP uses the same learned features to explain scores
across contexts, layers, and lenses. For a fixed token or contrast, the
score coefficients \(\alpha\) and feature coefficients
\(\beta_i(\alpha)\) stay fixed, so contributions change only with the
decoded state's alignment with each feature, measured by
\(p_i(\rstate)\) (Appendix~\ref{app:additional-margin-case-studies}).
Different tokens or contrasts can have different coefficients, and the
shared features let us compare which features support each reading.

The relative reconstruction error is
\(\rho_0(\rstate,\alpha)=|\epsilon|\,/\,|\sexact|\).
For a margin, agreement in sign between \(\sexact\) and \(\srecon\)
means both scores favor the same token. Coverage is the fraction of
reconstructions that preserve the sign of \(\sexact\) and have relative
error \(\rho_0\) below 50\%.
Appendix~\ref{app:score-decomposition} defines the adjusted relative
error used in some aggregate tables.

We also test whether the model makes similar predictions when its LM
head is replaced by \(\widehat{\WU}\), formed from the reconstructed
rows. Replacement fidelity measures how closely this head's token
probabilities and rankings match those of the original \(\WU\) on
held-out states (\S\ref{subsec:readout-replacement-fidelity}). We give
simpler decompositions of the readout rows the same tests to compare
their reconstruction accuracy with SRP's
(\S\ref{subsec:baseline-comparisons}).

\section{Evaluating SRP}
\label{sec:results}

We test how accurately SRP reconstructs scores, whether its feature
contributions distinguish word senses, and whether its token groups recur
across training runs. Appendix~\ref{app:causal-validation} tests
how closely individual contributions predict the score changes after we
remove their directions.

\paragraph{Models.}
We evaluate Qwen3.5~\citep{qwen35collection}, Gemma-4
\citep{gemma4official,gemma4collection}, Ministral~\citep{liu2026ministral3},
and Qwen/Llama readouts distilled for reasoning \citep{deepseekai2025deepseekr1}
with sizes from 0.8B to 9B parameters. Each of the eight models has a
separate readout SAE for evaluating reconstruction accuracy and replacement
fidelity (Appendix~\ref{app:robustness-controls}). The baseline comparison
covers the six readouts whose scores are linear in the decoded state
(\Cref{tab:readout-score-fidelity-summary}). We fixed this selection before
scoring and excluded the two Gemma-4 models because they apply a nonlinear
softcap after the linear readout
(Appendix~\ref{app:readout-factorizer-diagnostics-reproducibility}).
Table~\ref{tab:app-k-model-finalists} lists the models and dictionary settings.

\begin{table}[t]
\centering\footnotesize
\setlength{\tabcolsep}{2pt}
\renewcommand{\arraystretch}{1.06}
\textbf{A. Coverage across models \(\uparrow\)}\par\smallskip
\begin{tabular*}{\linewidth}{@{\extracolsep{\fill}}lccc@{}}
\toprule
Model & SRP & \makecell{Nearest-row\\ridge} & \makecell{k-means\\dictionary} \\
\midrule
Q-0.8B & \makecell{\textbf{75.4}\\{[73.3,77.3]}} & \makecell{66.5\\{[64.5,68.6]}} & \makecell{58.5\\{[56.6,60.3]}} \\
Q-2B & \makecell{\textbf{78.0}\\{[75.9,79.8]}} & \makecell{62.5\\{[60.6,64.4]}} & \makecell{59.6\\{[57.5,61.4]}} \\
Q-9B & \makecell{\textbf{81.2}\\{[79.2,83.1]}} & \makecell{63.9\\{[61.8,65.8]}} & \makecell{61.9\\{[59.8,63.9]}} \\
Min-8B & \makecell{\textbf{77.7}\\{[75.9,79.5]}} & \makecell{64.8\\{[62.9,66.5]}} & \makecell{62.8\\{[61.0,64.8]}} \\
R1-Q-7B & \makecell{\textbf{71.6}\\{[69.4,73.8]}} & \makecell{57.9\\{[56.0,59.8]}} & \makecell{47.9\\{[45.7,50.3]}} \\
R1-L-8B & \makecell{\textbf{75.1}\\{[72.8,77.2]}} & \makecell{60.0\\{[58.0,62.0]}} & \makecell{62.4\\{[60.6,64.2]}} \\
\bottomrule
\end{tabular*}
\par\medskip
\textbf{B. Pooled score reconstruction}\par\smallskip
\begin{tabular*}{\linewidth}{@{\extracolsep{\fill}}lcccc@{}}
\toprule
Method & \makecell{Med. rel.\\error (\%) \(\downarrow\)} & \makecell{Mean\\error \(\downarrow\)} & \makecell{95th\\pct. error \(\downarrow\)} & \makecell{Sign\\(\%) \(\uparrow\)} \\
\midrule
SRP & \textbf{10.1} & \textbf{0.69} & \textbf{2.34} & \textbf{92.6} \\
\makecell[l]{Nearest-row\\ridge} & 28.9 & 1.97 & 5.88 & 85.8 \\
\makecell[l]{k-means\\dictionary} & 36.4 & 2.62 & 7.34 & 83.9 \\
PCA-256 & 75.4 & 5.74 & 15.69 & 77.8 \\
\bottomrule
\end{tabular*}
\caption{\textbf{Fidelity of reconstructed logit differences.}
Panel A reports coverage in
percent with cluster-bootstrap 95\% CIs. Coverage requires matching sign
and absolute error below half the original score magnitude. Panel B
pools 8,009 contrasts across six models. Relative error has no denominator
floor, and mean and 95th-percentile absolute errors are in logits, pooled
across models with different logit scales. Ridge uses 128 neighbors, and
k-means uses 65,536 directions with 256 active coefficients per row.
Q/Min/R1-Q/R1-L \(=\) Qwen3.5 / Ministral-3 /
R1-Distill-Qwen / R1-Distill-Llama.
Appendix~\ref{app:robustness-controls} gives the metric definitions,
full baseline comparison, and model names.}
\label{tab:readout-score-fidelity-summary}
\end{table}

\subsection{How Faithful Are SRP's Reconstructions?}
\label{subsec:readout-replacement-fidelity}
\label{subsec:baseline-comparisons}
We test how accurately SRP reconstructs selected logit differences
before adding the residual. To measure the benefit of the learned
dictionary, we compare SRP with six simpler decompositions of the same
rows of \(\WU\). The baselines use ridge regression on neighbouring rows,
weighted kNN, k-means dictionaries at two widths, assignment to a single
cluster centroid, and PCA. For each model, every method reconstructs the
same scores (\S\ref{sec:method-reading-local-accounts}), for a total of
8{,}009 logit differences across six models. Each test case specifies an
input and an output comparison. We estimate confidence intervals by
resampling cases with replacement, keeping each case's rewordings and
context variants together
(Appendices~\ref{app:query-bank-provenance} and~\ref{app:direct-geometry-grid}).

On coverage, SRP leads the strongest alternative by 8.9--17.3 percentage
points across the six models, with nonoverlapping confidence intervals.
The advantage persists on Qwen3.5-2B when we match the k-means dictionary
to SRP in width and sparsity. Across the pooled scores, SRP also has the
lowest median relative error and the lowest mean and 95th-percentile
absolute errors
(\Cref{tab:readout-score-fidelity-summary}).
Coverage drops on Qwen3.5 and Gemma-4 when we shuffle codes or randomize
their support, showing that how features are assigned to rows matters
(Appendix~\ref{app:code-support-controls}).
For Qwen3.5-0.8B and Qwen3.5-2B, we also measure SRP's reconstruction
fidelity on contrasts derived from benchmarks
(\Cref{tab:benchmark-derived-readout-contrasts}).

To test fidelity across the vocabulary, we replace the original
\(\WU\) with its reconstruction and compare which token each LM head
ranks first on the same held-out states. Agreement varies across models
even when the reconstructions explain a similar share of the variance in
the readout rows. For example, the reconstructed Gemma-4-E2B head
captures most of the variance in the rows but preserves the top token
on only 33\% of states
(Appendix~\ref{app:readout-factorizer-transfer-comparisons}). On English C4
continuations \citep{raffel2020exploring,dodge2021documenting}, the
reconstructed Qwen3.5 heads preserve the top token more often when its
original margin over the runner-up is larger
(Appendix~\ref{app:readout-factorizer-transfer-comparisons}).

\subsection{Can SRP Distinguish Word Senses?}\label{subsec:same-token-analysis}\label{subsec:label-audit}\label{subsec:context-reweighting}
To illustrate one use of SRP, we test whether feature contributions can
distinguish different senses of the same token. The same LM-head row
scores \texttt{bug} in insect and software contexts. SRP shows how the
feature contributions to that score change with the prompt
(\S\ref{sec:method-comparing-accounts}).

\begin{figure}[t]
    \centering
    \includegraphics[width=\columnwidth]{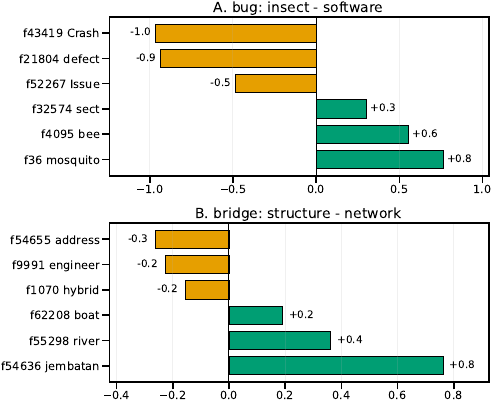}
    \caption{\textbf{Feature contributions to the same token across senses.}
    On Qwen3.5-2B (\(D=65536\), \(k=256\)), we score \texttt{bug} in insect
    and software contexts \textbf{(A)} and \texttt{bridge} in structure and
    network contexts \textbf{(B)}. Each bar gives the difference in one
    feature's contribution to the centered logit \(\rstateL^\top(w_v-\mu)\).
    Positive values indicate a larger contribution in the first context of
    each pair.}
    \label{fig:same-token-polysemy}
\end{figure}

On Qwen3.5-2B, insect-related features contribute more to the score for
\texttt{bug} in the insect context, and software-related features contribute
more in the software context. The score for \texttt{bridge} likewise draws
on different features in structure and network contexts
(\Cref{fig:same-token-polysemy}, Appendix~\ref{app:additional-margin-case-studies}).

\paragraph{Predicting senses in new contexts.}
To test whether these shifts generalize, we evaluate word-sense prediction
on CoarseWSD-20
\citep{loureiro2021coarsewsd}, a benchmark of contexts labelled with
the intended sense of a target word. For each prompt, we score the
known target word and use its SRP feature contributions to predict
its sense. To build a group for a sense, we
select the eight features whose mean contribution on the training
contexts differs most between that sense and the others.
We then assign each test context to the sense whose eight features add
up to the largest contribution. Selection uses training contexts alone,
and both selection and prediction use feature IDs and signed contributions,
without the descriptive labels.

\begin{table}[t]
\centering
\footnotesize
\setlength{\tabcolsep}{2pt}
\renewcommand{\arraystretch}{1.06}
\begin{tabular*}{\linewidth}{@{\extracolsep{\fill}}lccc@{}}
\toprule
 & \makecell{Qwen3.5\\2B} & \makecell{Qwen3.5\\9B} & \makecell{R1-Llama\\8B} \\
\midrule
Words evaluated & 20 & 20 & 13 \\
\midrule
Eight features per sense & 89.9 & 92.3 & 79.8 \\
Shuffled training labels & 40.8 & 40.7 & 42.2 \\
All SRP contributions & 94.7 & 97.0 & 89.2 \\
Full decoded state & 95.8 & 97.1 & 92.0 \\
\bottomrule
\end{tabular*}
\caption{\textbf{Word-sense prediction on held-out CoarseWSD-20 contexts.}
Results are balanced accuracies in percent. Each model is evaluated on
words that are single tokens in its vocabulary. The shuffled-label
control repeats feature selection with shuffled training sense labels.
The full-state reference uses a nearest centroid probe.}
\label{tab:sense-alignment-summary}
\end{table}

Every word exceeds its majority baseline at each tested group size
(Appendix~\ref{app:sense-labelled-evaluation}). Across all three models,
eight features per sense distinguish word senses in held-out contexts,
outperforming the shuffled-label control and retaining much of the
accuracy of the full set of SRP contributions
(\Cref{tab:sense-alignment-summary}).

\subsection{What Stays the Same Across Training Seeds?}
\label{subsec:stability}

We ask whether independently trained dictionaries recover similar token
groups for the same contrast. Prior work on activation SAEs motivates this
test by examining the stability of individual features across
initialization seeds while holding the training activations fixed
\citep{paulo2025stability}.
We also test how many of the same tokens simpler methods recover directly
from the unembedding rows.

On Qwen3.5-2B, we train three readout SAEs with different seeds at each
of two settings, \(32\times\) with \(k=256\) and \(16\times\) with \(k=128\)
(Appendix~\ref{app:basis-stability}). Each dictionary gives one group
of related vocabulary tokens for each of the two tokens in a contrast.
For \texttt{bug} versus \texttt{insect}, for example, we take the features
whose coefficients most favor \texttt{bug} and collect their top vocabulary
tokens, then do the same for \texttt{insect}. For each pair of dictionaries,
we use the tokens they both find for the same target as the reference.
We measure how many of those tokens the third dictionary, weighted
nearest neighbors, and hard clustering recover.

\begin{table}[t]
\centering
\footnotesize
\setlength{\tabcolsep}{4pt}
\renewcommand{\arraystretch}{1.08}
\begin{tabularx}{\linewidth}{@{}>{\RaggedRight\arraybackslash}Xcc@{}}
\toprule
Method & \(32\times\) & \(16\times\) \\
\midrule
Held-out SRP \mbox{(same recipe)} & 0.749 [.73, .77] & 0.720 [.70, .74] \\
Weighted row-kNN & 0.493 [.44, .54] & 0.432 [.38, .48] \\
Hard clustering & 0.249 [.21, .29] & 0.206 [.17, .24] \\
\midrule
SRP/kNN ratio & 1.52 & 1.67 \\
Cross-contrast null & 0.014 & 0.018 \\
\bottomrule
\end{tabularx}
\caption{\textbf{Recovery of shared token groups across training runs.}
On Qwen3.5-2B, all methods are evaluated on the same tokens agreed on by
two training runs at each width. Entries give recall with 95\% bootstrap
intervals clustered by contrast where shown. The null uses unrelated
contrasts.}
\label{tab:app-loo-core-recovery}
\end{table}

With the training recipe fixed, a fresh SRP dictionary recovers
72--75\% of the reference tokens, compared with 43--49\% for weighted
nearest neighbors (\Cref{tab:app-loo-core-recovery}). The full groups
vary across runs (Appendix~\ref{app:basis-stability-explanations}).

\FloatBarrier

\section{The Fitting Corpus Changes Lens Predictions}
\label{sec:beyond-token}
\label{sec:readout-analyses}
\label{subsec:cross-lens}

\begin{figure*}[t]
    \centering
    \includegraphics[width=\textwidth]{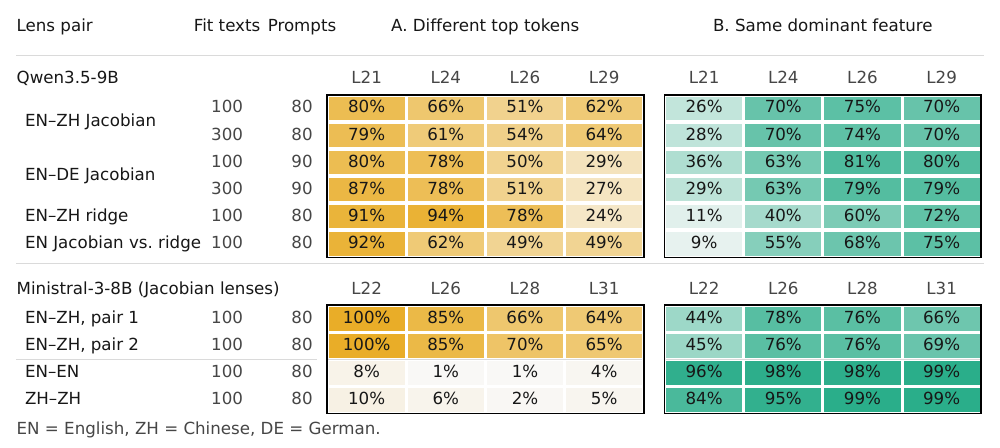}
    \caption{\textbf{Top-token disagreement and dominant-feature
    agreement by layer.} Each cell gives the percentage of evaluation
    prompts at that layer, with darker shading for higher percentages.
    Panel A compares the top tokens, using decoded strings on Qwen and
    token IDs on Ministral. Panel B compares the features with the largest
    absolute contribution to each lens's own top-token score. Both panels
    use all evaluation prompts. Fitting texts are counted per lens.
    On Ministral, top tokens differ much more often across fitting
    languages than between refits within a language. The same readout
    feature can dominate the scores of different top tokens. In every
    comparison across fitting languages, dominant features match more
    often at each later layer than at the earliest layer tested.}
    \label{fig:cross-lens-comparisons}
    \label{tab:cross-lens-token-disagreement}
    \label{tab:ministral-cross-lens}
    \label{tab:cross-lens-extensions}
\end{figure*}

A fitted lens maps an intermediate hidden state into the space used
by the LM head. The resulting tokens are often interpreted as evidence
about the information represented at that layer, including the language
in which the model is operating.
\citet{gurnee2026workspace} obtain English tokens from intermediate
states for a Chinese prompt and interpret them as evidence that the
intermediate computation occurs in English.

On Qwen3.5-9B and Ministral-3-8B, we find that changing the corpus used to
fit a lens can change its token predictions for the same hidden state.
We call this dependence \emph{corpus conditionality}. For instance, on
a Chinese prompt asking for the capital of the UK, a Jacobian lens
fitted on English predicts \texttt{London} and one fitted on Chinese
predicts \raisebox{-.2ex}{%
    \includegraphics[height=1.75ex]{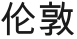}}.
Both lenses read the same Qwen3.5-9B state at layer 26.

\subsection{Does the Fitting Corpus Change the Top Token?}
\label{subsec:cross-lens-design}
\label{subsec:cross-lens-token-conflict}

\paragraph{Setup.}
In the main English--Chinese comparison, we average the Jacobian for each
lens over 100 C4 texts in its fitting language. The evaluation uses the same
80 prompts for every fit and holds the hidden states \(h_\ell\) fixed across
fits within each model. The lenses read the final prompt position at layers
21, 24, 26, and 29 on Qwen3.5-9B and at layers 22, 26, 28, and 31 on
Ministral-3-8B. On Qwen3.5-9B, we use the reference implementation
\citep{gurnee2026workspace}, with one fit per language and the same seed
and procedure. On Ministral-3-8B, two disjoint fitting sets per language,
each with 100 texts of 128 tokens, show how often predictions change when
the fitting texts are resampled within a language.

\paragraph{Results.}
On Ministral, the English and Chinese fits predict different top tokens
on 64--100\% of prompts across the tested layers, compared with 1--10\%
between refits within a language (\Cref{fig:cross-lens-comparisons}A).
At every tested layer, the disagreement across fitting languages is
larger than the variation we observe when resampling texts within a
language.

On Qwen, the English and Chinese fits rank different tokens first on
51--80\% of prompts across the tested layers. Panel A shows that
disagreement also occurs for English and German, two languages that
share a script, after we exclude cognates. For both language pairs,
disagreement rates are similar with 100 and 300 fitting texts. The
English and Chinese ridge translators related to the tuned lens
\citep{belrose2023tuned} also disagree. With the English fitting corpus held
fixed, the Jacobian lens and the ridge translator can still rank different
tokens first (\Cref{fig:cross-lens-comparisons}A).

To see whether these token changes affect the apparent language of the
prediction, we count the most frequent script across layers for each
prompt. On 39 of 80 Qwen prompts, that script is Latin for the English fit
and Chinese for the Chinese fit. The split is less frequent with 300
fitting texts, and the digit and proper noun controls, whose expected
tokens are the same in both languages, show no such split. The fitting
corpus can therefore change the apparent language of a lens prediction
from the same hidden state.

\subsection{What Does SRP Reveal When Lens Predictions Differ?}
\label{subsec:cross-lens-feature-invariance}
\label{subsec:cross-lens-extensions}

\paragraph{Setup.}
\label{subsec:cross-lens-actual-predictions}
For each model, we learn the dictionary from the unembedding rows
independently of the lens-fitting corpus and reuse it across fits
(\S\ref{sec:method-comparing-accounts}).
The same feature therefore represents the same readout direction in both
decompositions. The primary dictionary uses \(k=128\) on Qwen and
32\(\times\) expansion with \(k=256\) on Ministral.
For each lens, we use SRP to decompose its top token's score. We call the
feature with the largest absolute contribution the dominant feature.

\begin{figure}[t]
    \centering
    \includegraphics[width=\columnwidth]{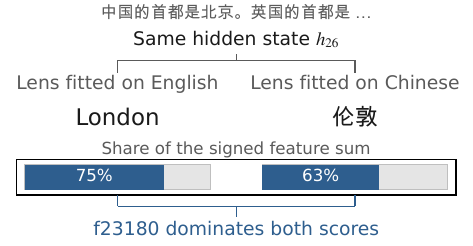}
    \caption{\textbf{Two token predictions share a dominant readout
    feature.} The Chinese prompt above translates to ``The capital of
    China is Beijing. The capital of the UK is \ldots''.
    The Jacobian lens fitted on English predicts \texttt{London}, and the
    one fitted on Chinese predicts \raisebox{-.2ex}{%
        \includegraphics[height=1.75ex]{figures/cross_lens/london_zh.pdf}}.
    Both read the same Qwen3.5-9B state at layer 26. The bars show feature
    f23180's share of each signed feature sum, which excludes the offset
    and residual.}
    \label{fig:cross-lens-butterfly}
\end{figure}

\paragraph{Results.}
At Qwen layer 26, the English and Chinese fits using 100 texts per lens
predict different top tokens on 51\% of prompts, and SRP finds the same
dominant feature on 75\% of prompts
(\Cref{fig:cross-lens-comparisons}A and B). Both percentages use all 80
prompts at that layer. The London example shows a case where both happen
on the same prompt (\Cref{fig:cross-lens-butterfly}).

To check how often the London pattern occurs, we restrict the
English--Chinese Jacobian comparisons to prompt--layer pairs with
different top tokens. SRP finds the same dominant feature in 44\% of
Qwen pairs and 60--61\% of Ministral pairs. On those Qwen pairs, mean
cosine similarity between signed contribution profiles is 0.54 and
falls to 0.44 after we remove the dominant features of both scores
from both profiles. The overlap therefore extends to other features.

Panel B shows how feature agreement varies across layers. In each
comparison across fitting languages, dominant features match more
often at every later tested layer than at the earliest one.
Appendix~\ref{app:all-layer-srp-lens} applies SRP to the ordinary logit lens
of Qwen3.5-2B at every layer of two prompts.
On Ministral, matches are more frequent between refits within a language
(\Cref{fig:cross-lens-comparisons}B).

The same refits also have more similar full contribution profiles. Over all
prompt--layer pairs, mean cosine similarity reaches 0.98--0.99 between
Ministral refits within a language and is about 0.70 for English--Chinese
comparisons on both models. The cross-language mean remains above 0.50
after we remove the dominant features. For comparison, pairing different
concepts from the same task family gives a mean full-profile cosine below
0.10. On Ministral, the \(16\times\), \(k=128\) dictionary gives similar
full-profile overlap across languages.
SRP identifies readout features shared by different token predictions. The greater
profile similarity between refits within a language on Ministral shows
that the fitting corpus also shapes how much those features contribute.

\section{Future Work}

\paragraph{Safety audits across lens fits.}
A safety audit could test how sensitive inferences about evaluation awareness
are to the lens fitting corpus. With language and task held fixed, we could
vary how often the fitting corpus discusses oversight and compare signed
contributions to the same action contrast on identical hidden states.
Separately varying oversight cues would test whether those contributions
track changes in the model's actual choices.

\paragraph{Layer attribution and causal tests.}
In the appendix, we provide an initial exploration by combining SRP with
direct logit attribution to break down each layer's contribution to the
\texttt{verify} minus \texttt{assume} margin by readout feature. The same
analysis could identify layers that contribute to features shared across
languages. Activation patching could then test whether those layers causally
support equivalent answers.

\paragraph{Larger models and realistic tasks.}
We could apply SRP to larger open-weight models in multi-turn tasks with
tool use, building on Jacobian lens analyses of audit transcripts
\citep{gurnee2026workspace}. Decomposing scores for competing actions would
let us trace how support for each action develops over an interaction.

\paragraph{Low-rank readout factorization.}
We could adapt the activation-aware factorization of
\citet{conf/iclr/01200W025} to the LM head, fitting low-rank factors to
preserve logits on a fixed set of decoded states. The resulting factors
would provide an alternative basis for decomposing lens scores.

\section{Conclusion}

A fitted lens can rank different tokens first for the same hidden state
when only its fitting corpus changes. Sparse Readout Prism offers shared
readout features as a unit of interpretation for lens scores. It learns
these features from the unembedding and measures their signed
contributions to each score. Across six models, SRP reconstructs logit
differences with higher coverage than the strongest geometric baseline
for each model. Comparing feature contributions distinguishes senses of a
fixed token in held-out contexts on all three models tested. The same
readout features can support different top tokens across fits, including
tokens in different languages. In these cases, SRP offers an alternative
way to interpret lens predictions with less dependence on the fitting
corpus.

\section*{Limitations}

\paragraph{Fitting cost.}
SRP requires fitting a sparse autoencoder once per model, taking roughly
6--12 GPU-hours on one NVIDIA A40 in our experiments, and the fitted
dictionary is then reused across layers and composed with different
lenses. Fitting the readout rows serves to give a basis independent of
the lens-fitting corpus, and we make no claim about compute savings. For
context, fitting one Jacobian lens on 100 prompts for Qwen3.5-9B took
about 11 A40 GPU-hours, and the recipe of \citet{gurnee2026workspace}
uses about 1{,}000 prompts, which a linear estimate puts near 110
GPU-hours.

\paragraph{Feature identity across fits.}
As with activation SAEs, the decoder directions learned by SRP depend on
dictionary size, sparsity \(k\), training recipe, and seed
\citep{elhage2022superposition,karvonen2025saebench,makelov2024principled,paulo2025stability}.
The tokens associated with one feature in one fit can be spread across
several features in another. In our Qwen3.5-2B seed study, allowing up to
three matches recovers a median of 83--90\% of a feature's top tokens across
pairs of fits (\S\ref{subsec:stability}). Individual features therefore
define units of attribution within a fitted dictionary. Two dictionaries
can reconstruct the same score and allocate its contributions differently.

\section*{Ethical Considerations}

This work studies the readouts of pretrained language models, collects no
new data from human subjects, and evaluates no deployed systems. Experiments use public
checkpoints and held-out decoded states computed from text that
includes C4 continuations, which may contain offensive, private,
copyrighted, or sensitive material
\citep{raffel2020exploring,dodge2021documenting}. We use C4 to measure
readout reconstruction and fit the lenses, and make no claims about the social validity of
individual tokens, labels, or generated outputs. Checkpoints and
the C4 data are used under the licenses or terms published on their
release pages and model cards. We do not redistribute the
checkpoints or raw C4 text. The released SRP dictionaries%
\ifshowidentity\ (\href{https://huggingface.co/hematteo/sparse-readout-prism}{\texttt{hematteo/sparse-readout-prism}})\fi\
carry license and usage notes specific to each artifact.

SRP makes lexical readout directions and row labels easier to inspect.
Evaluation of safety filters, moderation, and deployment controls raises
questions at the system level that lie beyond this work.

\bibliography{references}

\begin{thebibliography}{42}
\providecommand{\natexlab}[1]{#1}

\bibitem[{Arora et~al.(2018)Arora, Li, Liang, Ma, and
  Risteski}]{arora2018polysemy}
Sanjeev Arora, Yuanzhi Li, Yingyu Liang, Tengyu Ma, and Andrej Risteski. 2018.
\newblock \href {https://doi.org/10.1162/tacl_a_00034} {Linear algebraic
  structure of word senses, with applications to polysemy}.
\newblock \emph{Transactions of the Association for Computational Linguistics},
  6:483--495.

\bibitem[{Belrose et~al.(2023)Belrose, Ostrovsky, McKinney, Furman, Smith,
  Halawi, Biderman, and Steinhardt}]{belrose2023tuned}
Nora Belrose, Igor Ostrovsky, Lev McKinney, Zach Furman, Logan Smith, Danny
  Halawi, Stella Biderman, and Jacob Steinhardt. 2023.
\newblock \href {https://arxiv.org/abs/2303.08112} {Eliciting latent
  predictions from transformers with the tuned lens}.
\newblock \emph{Preprint}, arXiv:2303.08112.

\bibitem[{Braun et~al.(2025)Braun, Bushnaq, Heimersheim, Mendel, and
  Sharkey}]{braun2025parameter}
Dan Braun, Lucius Bushnaq, Stefan Heimersheim, Jake Mendel, and Lee Sharkey.
  2025.
\newblock \href {https://arxiv.org/abs/2501.14926} {Interpretability in
  parameter space: Minimizing mechanistic description length with
  attribution-based parameter decomposition}.
\newblock \emph{Preprint}, arXiv:2501.14926.

\bibitem[{Bricken et~al.(2023)Bricken, Templeton, Batson, Chen, Jermyn,
  Conerly, Turner, Anil, Denison, Askell, Lasenby, Wu, Kravec, Schiefer,
  Maxwell, Joseph, Hatfield-Dodds, Tamkin, Nguyen, McLean, Burke, Hume, Carter,
  Henighan, and Olah}]{bricken2023monosemanticity}
Trenton Bricken, Adly Templeton, Joshua Batson, Brian Chen, Adam Jermyn, Tom
  Conerly, Nicholas~L. Turner, Cem Anil, Carson Denison, Amanda Askell, Robert
  Lasenby, Yifan Wu, Shauna Kravec, Nicholas Schiefer, Tim Maxwell, Nicholas
  Joseph, Zac Hatfield-Dodds, Alex Tamkin, Karina Nguyen, and 6 others. 2023.
\newblock \href
  {https://transformer-circuits.pub/2023/monosemantic-features/index.html}
  {Towards monosemanticity: Decomposing language models with dictionary
  learning}.
\newblock \emph{Transformer Circuits Thread}.

\bibitem[{{DeepSeek-AI}(2025)}]{deepseekai2025deepseekr1}
{DeepSeek-AI}. 2025.
\newblock \href {https://doi.org/10.1038/s41586-025-09422-z} {{DeepSeek-R1}
  incentivizes reasoning in {LLM}s through reinforcement learning}.
\newblock \emph{Nature}, 645(8081):633--638.

\bibitem[{Dodge et~al.(2021)Dodge, Sap, Marasovi{\'c}, Agnew, Ilharco,
  Groeneveld, Mitchell, and Gardner}]{dodge2021documenting}
Jesse Dodge, Maarten Sap, Ana Marasovi{\'c}, William Agnew, Gabriel Ilharco,
  Dirk Groeneveld, Margaret Mitchell, and Matt Gardner. 2021.
\newblock \href {https://doi.org/10.18653/v1/2021.emnlp-main.98} {Documenting
  large webtext corpora: A case study on the colossal clean crawled corpus}.
\newblock In \emph{Proceedings of the 2021 Conference on Empirical Methods in
  Natural Language Processing}, pages 1286--1305.

\bibitem[{Elhage et~al.(2022)Elhage, Hume, Olsson, Schiefer, Henighan, Kravec,
  Hatfield-Dodds, Lasenby, Drain, Chen, Grosse, McCandlish, Kaplan, Amodei,
  Wattenberg, and Olah}]{elhage2022superposition}
Nelson Elhage, Tristan Hume, Catherine Olsson, Nicholas Schiefer, Tom Henighan,
  Shauna Kravec, Zac Hatfield-Dodds, Robert Lasenby, Dawn Drain, Carol Chen,
  Roger Grosse, Sam McCandlish, Jared Kaplan, Dario Amodei, Martin Wattenberg,
  and Christopher Olah. 2022.
\newblock \href {https://transformer-circuits.pub/2022/toy_model/index.html}
  {Toy models of superposition}.
\newblock \emph{Transformer Circuits Thread}.

\bibitem[{Elhage et~al.(2021)Elhage, Nanda, Olsson, Henighan, Joseph, Mann,
  Askell, Bai, Chen, Conerly, DasSarma, Drain, Ganguli, Hatfield-Dodds,
  Hernandez, Jones, Kernion, Lovitt, Ndousse, Amodei, Brown, Clark, Kaplan,
  McCandlish, and Olah}]{elhage2021framework}
Nelson Elhage, Neel Nanda, Catherine Olsson, Tom Henighan, Nicholas Joseph, Ben
  Mann, Amanda Askell, Yuntao Bai, Anna Chen, Tom Conerly, Nova DasSarma, Dawn
  Drain, Deep Ganguli, Zac Hatfield-Dodds, Danny Hernandez, Andy Jones, Jackson
  Kernion, Liane Lovitt, Kamal Ndousse, and 6 others. 2021.
\newblock \href {https://transformer-circuits.pub/2021/framework/index.html} {A
  mathematical framework for transformer circuits}.
\newblock \emph{Transformer Circuits Thread}.

\bibitem[{Gao et~al.(2025)Gao, Dupr{\'e}~la Tour, Tillman, Goh, Troll, Radford,
  Sutskever, Leike, and Wu}]{gao2024scaling}
Leo Gao, Tom Dupr{\'e}~la Tour, Henk Tillman, Gabriel Goh, Rajan Troll, Alec
  Radford, Ilya Sutskever, Jan Leike, and Jeffrey Wu. 2025.
\newblock \href {https://arxiv.org/abs/2406.04093} {Scaling and evaluating
  sparse autoencoders}.
\newblock In \emph{International Conference on Learning Representations}.

\bibitem[{Ghandeharioun et~al.(2024)Ghandeharioun, Caciularu, Pearce, Dixon,
  and Geva}]{ghandeharioun2024patchscopes}
Asma Ghandeharioun, Avi Caciularu, Adam Pearce, Lucas Dixon, and Mor Geva.
  2024.
\newblock \href {https://arxiv.org/abs/2401.06102} {Patchscopes: A unifying
  framework for inspecting hidden representations of language models}.
\newblock In \emph{Proceedings of the 41st International Conference on Machine
  Learning}, volume 235 of \emph{Proceedings of Machine Learning Research},
  pages 15466--15490. PMLR.

\bibitem[{{Google DeepMind}(2026{\natexlab{a}})}]{gemma4official}
{Google DeepMind}. 2026{\natexlab{a}}.
\newblock {Gemma 4}.
\newblock \url{https://deepmind.google/models/gemma/gemma-4/}.
\newblock Official model-family page; accessed 2026-05-20.

\bibitem[{{Google DeepMind}(2026{\natexlab{b}})}]{gemma4collection}
{Google DeepMind}. 2026{\natexlab{b}}.
\newblock {Gemma 4} model collection.
\newblock \url{https://huggingface.co/collections/google/gemma-4}.
\newblock Hugging Face model collection; accessed 2026-05-20.

\bibitem[{Guha et~al.(2023)Guha, Nyarko, Ho, R{\'e}, Chilton, Narayana,
  Chohlas-Wood, Peters, Waldon, Rockmore, Zambrano, Talisman, Hoque, Surani,
  Fagan, Sarfaty, Dickinson, Porat, Hegland, Wu, Nudell, Niklaus, Nay, Choi,
  Tobia, Hagan, Ma, Livermore, Rasumov-Rahe, Holzenberger, Kolt, Henderson,
  Rehaag, Goel, Gao, Williams, Gandhi, Zur, Iyer, and Li}]{guha2023legalbench}
Neel Guha, Julian Nyarko, Daniel~E. Ho, Christopher R{\'e}, Adam Chilton,
  Aditya Narayana, Alex Chohlas-Wood, Austin Peters, Brandon Waldon, Daniel~N.
  Rockmore, Diego Zambrano, Dmitry Talisman, Enam Hoque, Faiz Surani, Frank
  Fagan, Galit Sarfaty, Gregory~M. Dickinson, Haggai Porat, Jason Hegland, and
  21 others. 2023.
\newblock \href {https://arxiv.org/abs/2308.11462} {{LegalBench}: A
  collaboratively built benchmark for measuring legal reasoning in large
  language models}.
\newblock In \emph{Advances in Neural Information Processing Systems
  ({NeurIPS}) Datasets and Benchmarks Track}, volume~36.

\bibitem[{Gurnee et~al.(2026)Gurnee, Sofroniew, Pearce, Piotrowski, Kauvar,
  Chen, Soligo, Bogdan, Ong, Wang, Thompson, Abrahams, Kantamneni, Ameisen,
  Batson, and Lindsey}]{gurnee2026workspace}
Wes Gurnee, Nicholas Sofroniew, Adam Pearce, Mateusz Piotrowski, Isaac Kauvar,
  Runjin Chen, Anna Soligo, Paul Bogdan, Euan Ong, Rowan Wang, T.~Ben Thompson,
  David Abrahams, Subhash Kantamneni, Emmanuel Ameisen, Joshua Batson, and Jack
  Lindsey. 2026.
\newblock \href {https://transformer-circuits.pub/2026/workspace/}
  {Verbalizable representations form a global workspace in language models}.
\newblock \emph{Transformer Circuits Thread}.

\bibitem[{He et~al.(2026)He, Shen, Iacob, Jovanovic, Qiu, and
  Lane}]{he2026learningreadoutunembedding}
Matteo He, William~F. Shen, Alex Iacob, Andrej Jovanovic, Xinchi Qiu, and
  Nicholas~D. Lane. 2026.
\newblock \href {https://arxiv.org/abs/2610.21025} {Learning to read out:
  Unembedding dynamics in language model pretraining}.
\newblock \emph{Accepted at NeurIPS 2026}, arXiv:2610.21025.

\bibitem[{Huben et~al.(2024)Huben, Cunningham, Smith, Ewart, and
  Sharkey}]{cunningham2023sparse}
Robert Huben, Hoagy Cunningham, Logan Smith, Aidan Ewart, and Lee Sharkey.
  2024.
\newblock \href {https://arxiv.org/abs/2309.08600} {Sparse autoencoders find
  highly interpretable features in language models}.
\newblock In \emph{International Conference on Learning Representations}.

\bibitem[{Karvonen et~al.(2025)Karvonen, Rager, Lin, Tigges, Bloom, Chanin,
  Lau, Farrell, Mcdougall, Ayonrinde, Till, Wearden, Conmy, Marks, and
  Nanda}]{karvonen2025saebench}
Adam Karvonen, Can Rager, Johnny Lin, Curt Tigges, Joseph~Isaac Bloom, David
  Chanin, Yeu-Tong Lau, Eoin Farrell, Callum~Stuart Mcdougall, Kola Ayonrinde,
  Demian Till, Matthew Wearden, Arthur Conmy, Samuel Marks, and Neel Nanda.
  2025.
\newblock \href {https://arxiv.org/abs/2503.09532} {{SAEBench}: A comprehensive
  benchmark for sparse autoencoders in language model interpretability}.
\newblock In \emph{Proceedings of the 42nd International Conference on Machine
  Learning}, volume 267 of \emph{Proceedings of Machine Learning Research},
  pages 29223--29264. PMLR.

\bibitem[{Koreeda and Manning(2021)}]{koreeda2021contractnli}
Yuta Koreeda and Christopher~D. Manning. 2021.
\newblock \href {https://doi.org/10.18653/v1/2021.findings-emnlp.164}
  {{ContractNLI}: A dataset for document-level natural language inference for
  contracts}.
\newblock In \emph{Findings of the Association for Computational Linguistics:
  EMNLP 2021}, pages 1907--1919.

\bibitem[{Lindsey et~al.(2024)Lindsey, Templeton, Marcus, Conerly, Batson, and
  Olah}]{lindsey2024crosscoders}
Jack Lindsey, Adly Templeton, Jonathan Marcus, Thomas Conerly, Joshua Batson,
  and Christopher Olah. 2024.
\newblock \href {https://transformer-circuits.pub/2024/crosscoders/index.html}
  {Sparse crosscoders for cross-layer features and model diffing}.
\newblock \emph{Transformer Circuits Thread}.

\bibitem[{Liu and Deng(2026)}]{liu2026weightbasedsae}
Yiting Liu and Zhi-Hong Deng. 2026.
\newblock \href {https://arxiv.org/abs/2601.22447} {Beyond activation patterns:
  A weight-based out-of-context explanation of sparse autoencoder features}.
\newblock \emph{Preprint}, arXiv:2601.22447.

\bibitem[{{Llama Team}(2024)}]{grattafiori2024llama3herd}
{Llama Team}. 2024.
\newblock \href {https://arxiv.org/abs/2407.21783} {The {Llama 3} herd of
  models}.
\newblock \emph{Preprint}, arXiv:2407.21783.

\bibitem[{Loureiro et~al.(2021)Loureiro, Rezaee, Pilehvar, and
  Camacho-Collados}]{loureiro2021coarsewsd}
Daniel Loureiro, Kiamehr Rezaee, Mohammad~Taher Pilehvar, and Jose
  Camacho-Collados. 2021.
\newblock \href {https://doi.org/10.1162/coli_a_00405} {Analysis and evaluation
  of language models for word sense disambiguation}.
\newblock \emph{Computational Linguistics}, 47(2):387--443.

\bibitem[{Makelov et~al.(2025)Makelov, Lange, and
  Nanda}]{makelov2024principled}
Aleksandar Makelov, Georg Lange, and Neel Nanda. 2025.
\newblock \href {https://arxiv.org/abs/2405.08366} {Towards principled
  evaluations of sparse autoencoders for interpretability and control}.
\newblock In \emph{International Conference on Learning Representations
  ({ICLR})}.

\bibitem[{Makhzani and Frey(2014)}]{makhzani2013ksparse}
Alireza Makhzani and Brendan Frey. 2014.
\newblock \href {https://arxiv.org/abs/1312.5663} {k-sparse autoencoders}.
\newblock In \emph{International Conference on Learning Representations
  ({ICLR})}.

\bibitem[{Merity et~al.(2017)Merity, Xiong, Bradbury, and
  Socher}]{merity2017pointer}
Stephen Merity, Caiming Xiong, James Bradbury, and Richard Socher. 2017.
\newblock \href {https://arxiv.org/abs/1609.07843} {Pointer sentinel mixture
  models}.
\newblock In \emph{5th International Conference on Learning Representations}.

\bibitem[{{Mistral AI}(2026)}]{liu2026ministral3}
{Mistral AI}. 2026.
\newblock \href {https://arxiv.org/abs/2601.08584} {{Ministral 3}}.
\newblock \emph{Preprint}, arXiv:2601.08584.

\bibitem[{Nguyen(2024)}]{nguyen2024logitprisms}
Thong~T. Nguyen. 2024.
\newblock Logit prisms: Decomposing transformer outputs for mechanistic
  interpretability.
\newblock \url{https://neuralblog.github.io/logit-prisms/}.

\bibitem[{nostalgebraist(2020)}]{nostalgebraist2020}
nostalgebraist. 2020.
\newblock Interpreting {GPT}: The logit lens.
\newblock
  \url{https://www.lesswrong.com/posts/AcKRB8wDpdaN6v6ru/interpreting-gpt-the-logit-lens}.

\bibitem[{Pal et~al.(2023)Pal, Sun, Yuan, Wallace, and Bau}]{pal2023future}
Koyena Pal, Jiuding Sun, Andrew Yuan, Byron~C. Wallace, and David Bau. 2023.
\newblock \href {https://doi.org/10.18653/v1/2023.conll-1.37} {Future lens:
  Anticipating subsequent tokens from a single hidden state}.
\newblock In \emph{Proceedings of the 27th Conference on Computational Natural
  Language Learning (CoNLL)}, pages 548--560.

\bibitem[{Paulo and Belrose(2026)}]{paulo2025stability}
Gon\c{c}alo Paulo and Nora Belrose. 2026.
\newblock \href {https://arxiv.org/abs/2501.16615} {Sparse autoencoders trained
  on the same data learn different features}.
\newblock In \emph{International Conference on Learning Representations
  ({ICLR})}.

\bibitem[{{Qwen Team}(2026)}]{qwen35collection}
{Qwen Team}. 2026.
\newblock {Qwen3.5} model collection.
\newblock \url{https://huggingface.co/collections/Qwen/qwen35}.
\newblock Hugging Face model collection; accessed 2026-05-20.

\bibitem[{Raffel et~al.(2020)Raffel, Shazeer, Roberts, Lee, Narang, Matena,
  Zhou, Li, and Liu}]{raffel2020exploring}
Colin Raffel, Noam Shazeer, Adam Roberts, Katherine Lee, Sharan Narang, Michael
  Matena, Yanqi Zhou, Wei Li, and Peter~J. Liu. 2020.
\newblock \href {https://jmlr.org/papers/v21/20-074.html} {Exploring the limits
  of transfer learning with a unified text-to-text transformer}.
\newblock \emph{Journal of Machine Learning Research}, 21(140):1--67.

\bibitem[{Rajpurkar et~al.(2018)Rajpurkar, Jia, and
  Liang}]{rajpurkar2018squad2}
Pranav Rajpurkar, Robin Jia, and Percy Liang. 2018.
\newblock \href {https://doi.org/10.18653/v1/P18-2124} {Know what you don't
  know: Unanswerable questions for {SQuAD}}.
\newblock In \emph{Proceedings of the 56th Annual Meeting of the Association
  for Computational Linguistics (Volume 2: Short Papers)}, pages 784--789.

\bibitem[{Schut et~al.(2025)Schut, Gal, and Farquhar}]{schut2025english}
Lisa Schut, Yarin Gal, and Sebastian Farquhar. 2025.
\newblock \href {https://arxiv.org/abs/2502.15603} {Do multilingual {LLM}s
  think in {English}?}
\newblock \emph{Preprint}, arXiv:2502.15603.

\bibitem[{Siddiq and Santos(2022)}]{siddiq2022securityeval}
Mohammed~Latif Siddiq and Joanna C.~S. Santos. 2022.
\newblock \href {https://doi.org/10.1145/3549035.3561184} {{SecurityEval}
  dataset: Mining vulnerability examples to evaluate machine learning-based
  code generation techniques}.
\newblock In \emph{Proceedings of the 1st International Workshop on Mining
  Software Repositories Applications for Privacy and Security}, pages 29--33.

\bibitem[{Subramanian et~al.(2018)Subramanian, Pruthi, Jhamtani,
  Berg-Kirkpatrick, and Hovy}]{subramanian2018spine}
Anant Subramanian, Danish Pruthi, Harsh Jhamtani, Taylor Berg-Kirkpatrick, and
  Eduard Hovy. 2018.
\newblock \href {https://doi.org/10.1609/aaai.v32i1.11935} {{SPINE}: {SP}arse
  {I}nterpretable {N}eural {E}mbeddings}.
\newblock In \emph{Proceedings of the Thirty-Second {AAAI} Conference on
  Artificial Intelligence}, pages 4921--4928, New Orleans, Louisiana, USA. AAAI
  Press.

\bibitem[{Templeton et~al.(2024)Templeton, Conerly, Marcus, Lindsey, Bricken,
  Chen, Pearce, Citro, Ameisen, Jones, Cunningham, Turner, McDougall,
  MacDiarmid, Tamkin, Durmus, Hume, Mosconi, Freeman, Sumers, Rees, Batson,
  Jermyn, Carter, Olah, and Henighan}]{templeton2024scaling}
Adly Templeton, Tom Conerly, Jonathan Marcus, Jack Lindsey, Trenton Bricken,
  Brian Chen, Adam Pearce, Craig Citro, Emmanuel Ameisen, Andy Jones, Hoagy
  Cunningham, Nicholas~L. Turner, Callum McDougall, Monte MacDiarmid, Alex
  Tamkin, Esin Durmus, Tristan Hume, Francesco Mosconi, C.~Daniel Freeman, and
  7 others. 2024.
\newblock \href
  {https://transformer-circuits.pub/2024/scaling-monosemanticity/index.html}
  {Scaling monosemanticity: Extracting interpretable features from {Claude} 3
  {Sonnet}}.
\newblock \emph{Transformer Circuits Thread}.

\bibitem[{Wang et~al.(2023)Wang, Variengien, Conmy, Shlegeris, and
  Steinhardt}]{wang2022ioi}
Kevin Wang, Alexandre Variengien, Arthur Conmy, Buck Shlegeris, and Jacob
  Steinhardt. 2023.
\newblock \href {https://arxiv.org/abs/2211.00593} {Interpretability in the
  wild: A circuit for indirect object identification in {GPT-2} small}.
\newblock In \emph{International Conference on Learning Representations
  ({ICLR})}.

\bibitem[{Wang et~al.(2025)Wang, Zheng, Wan, and Zhang}]{conf/iclr/01200W025}
Xin Wang, Yu~Zheng, Zhongwei Wan, and Mi~Zhang. 2025.
\newblock \href
  {https://proceedings.iclr.cc/paper_files/paper/2025/hash/3104e1ab39875cf54fe1eb4473e7c5a1-Abstract-Conference.html}
  {{SVD-LLM}: Truncation-aware singular value decomposition for large language
  model compression}.
\newblock In \emph{The Thirteenth International Conference on Learning
  Representations}.

\bibitem[{Wendler et~al.(2024)Wendler, Veselovsky, Monea, and
  West}]{wendler2024llamas}
Chris Wendler, Veniamin Veselovsky, Giovanni Monea, and Robert West. 2024.
\newblock \href {https://doi.org/10.18653/v1/2024.acl-long.820} {Do llamas work
  in {English}? {O}n the latent language of multilingual transformers}.
\newblock In \emph{Proceedings of the 62nd Annual Meeting of the Association
  for Computational Linguistics (Volume 1: Long Papers)}, pages 15366--15394.

\bibitem[{Yang et~al.(2024)Yang, Zhang, Hui, Gao, Yu, Li, Liu, Tu, Zhou, Lin,
  Lu, Xue, Lin, Liu, Ren, and Zhang}]{yang2024qwen25math}
An~Yang, Beichen Zhang, Binyuan Hui, Bofei Gao, Bowen Yu, Chengpeng Li,
  Dayiheng Liu, Jianhong Tu, Jingren Zhou, Junyang Lin, Keming Lu, Mingfeng
  Xue, Runji Lin, Tianyu Liu, Xingzhang Ren, and Zhenru Zhang. 2024.
\newblock \href {https://arxiv.org/abs/2409.12122} {{Qwen2.5-Math} technical
  report: Toward mathematical expert model via self-improvement}.
\newblock \emph{Preprint}, arXiv:2409.12122.

\bibitem[{Yang et~al.(2018)Yang, Qi, Zhang, Bengio, Cohen, Salakhutdinov, and
  Manning}]{yang2018hotpotqa}
Zhilin Yang, Peng Qi, Saizheng Zhang, Yoshua Bengio, William~W. Cohen, Ruslan
  Salakhutdinov, and Christopher~D. Manning. 2018.
\newblock \href {https://doi.org/10.18653/v1/D18-1259} {{HotpotQA}: A dataset
  for diverse, explainable multi-hop question answering}.
\newblock In \emph{Proceedings of the 2018 Conference on Empirical Methods in
  Natural Language Processing}, pages 2369--2380.

\end{thebibliography}

\clearpage
\appendix
\AppendixFloatSetup
\section*{Appendix Roadmap}
\addcontentsline{toc}{section}{Appendix Roadmap}

\begingroup
\footnotesize
\setlength{\tabcolsep}{3pt}
\renewcommand{\arraystretch}{1.12}
\newcommand{\AppRoadmapEntry}[2]{%
  \textbf{\ref*{#1}} &
  \nameref{#1} \textnormal{(p.~\pageref{#1}).}
  \newline #2 \\
}
\noindent\begin{tabularx}{\linewidth}{@{}p{1.2em}Y@{}}
\toprule
 & \textbf{Contents} \\
\midrule
\AppRoadmapEntry{app:score-decomposition}
  {Score families and relative error conventions.}
\AppRoadmapEntry{app:model-suite-selection}
  {Training recipe, selection checks, and reproducibility.}
\AppRoadmapEntry{app:robustness-controls}
  {Head replacement, baselines, near ties, and evaluation prompts.}
\AppRoadmapEntry{app:additional-qwen-display-examples}
  {Context effects, feature labels, held-out senses, stability across runs, and the logit lens across layers.}
\AppRoadmapEntry{app:readout-interventions}
  {Predicted and measured score drops, including overlap and residual effects.}
\AppRoadmapEntry{app:feature-resolved-dla}
  {Margin contributions by feature and residual-stream component.}
\AppRoadmapEntry{app:lexical-editing-matched-kl}
  {Transfer to held-out tokens and comparisons at matched KL.}
\bottomrule
\end{tabularx}
\endgroup

% Score definitions and reconstruction error for Sparse Readout Prism.

\section{Score Definitions and Reconstruction Error}
\label{app:score-decomposition}
\label{app:interpreting-sparse-readout-prism-figures}
\label{app:additional-diagnostics}

\subsection{Score Families}

Table~\ref{tab:method-special-cases} gives the readout direction
\(q_\alpha\) and feature coefficients \(\beta_i(\alpha)\) for five score
families, using the decomposition in \S\ref{sec:method-local-score}.

\par\addvspace{\intextsep}
\noindent\begin{minipage}{\linewidth}
\centering
\footnotesize
\setlength{\tabcolsep}{3pt}
\renewcommand{\arraystretch}{1.14}
\begin{tabularx}{\linewidth}{@{}>{\RaggedRight\arraybackslash}p{0.34\linewidth}Z Z@{}}
\toprule
Selected score family & \(q_\alpha\) & \(\beta_i(\alpha)\) \\
\midrule
\textbf{Token logit} & \(w_A\) & \(z_{A,i}\) \\
\addlinespace[1pt]
\textbf{Vocabulary-mean contrast} & \(w_A-\bar{w}_{\mathcal{V}}\) &
\(z_{A,i}-\bar{z}_{\mathcal{V},i}\) \\
\addlinespace[1pt]
\textbf{Logit difference} & \(w_A-w_B\) & \(z_{A,i}-z_{B,i}\) \\
\addlinespace[1pt]
\textbf{Group contrast} &
\(\bar{w}_{\mathcal{A}}-\bar{w}_{\mathcal{B}}\) &
\(\bar{z}_{\mathcal{A},i}-\bar{z}_{\mathcal{B},i}\) \\
\addlinespace[1pt]
\textbf{Top token vs. selected competitors} &
\(w_u-\bar{w}_{\pi,\mathcal{R}}\) &
\(z_{u,i}-\bar{z}_{\pi,\mathcal{R},i}\) \\
\bottomrule
\end{tabularx}
\captionof{table}{\textbf{Readout directions and feature coefficients by score.}
Each score uses the same token weights for readout rows and feature
coefficients.
\(A,B\in\mathcal{V}\) are tokens, \(\mathcal{A},\mathcal{B}\subset\mathcal{V}\)
are token groups, and \(u\) is the original LM head's top token.
Bars denote uniform averages over the indicated set unless weighted by
\(\pi\). The weights \(\pi_r\) are nonnegative and sum to one over the
competitors \(\mathcal{R}\), with weight one for a single competitor.}
\label{tab:method-special-cases}
\end{minipage}
\par\addvspace{\intextsep}

\subsection{Reconstruction Error}
\label{app:score-decomposition-fidelity}
\label{app:display-taxonomy}
\label{app:display-residual-decomposition}

The relative reconstruction error with \(\delta\) logits added to the
denominator is
\[
    \rho_\delta(\rstate,\alpha)
    =\frac{|\sexact-\srecon|}
    {|\sexact|+\delta}.
\]
Case study figures, baseline comparisons, and identity checks use ordinary
relative error \(\rho_0\) from \S\ref{sec:method-reading-local-accounts}.
Tables that report \(\rho_{0.5}(\rstate,\alpha)\) add \(\delta=0.5\) logits
to each denominator so near ties have less influence on the median.

\section{Dictionary Training and Selection}
\label{app:model-suite-selection}
\label{sec:app-model-suite-selection}
\label{sec:app-model-suite-criteria}
\label{sec:app-model-suite-chosen}
\label{sec:app-model-suite-boundaries}

% Readout SAE training and evaluation details.

\subsection{Training and Evaluation Recipe}
\label{app:readout-factorizer-diagnostics-reproducibility}
\label{app:readout-factorizer-limitations}

\paragraph{Training data.}
For each model, the recipe collects LM head inputs from unlabelled English
passages in the WikiText-2 test split \citep{merity2017pointer}. If too few
passages meet the length requirement, the training script fills the
remaining slots with predefined prompts. The inputs are sampled
independently of the lens fits.

\paragraph{Row sampling.}
The collected LM head inputs supply the predictions used for row sampling
and the auxiliary loss. To set the sampling weights, the recipe counts
each token's appearances among the highest-ranked predictions,
applies a minimum count, and normalizes the counts. Training draws rows
with replacement from an equal mixture of uniform and frequency-weighted
sampling (\texttt{hybrid\_50freq\_50uniform}).

\paragraph{Auxiliary logit loss.}
The auxiliary loss in \S\ref{sec:method-factorization} directly penalizes
reconstruction error in the highest-ranked token's logit at each position
(\texttt{prism\_top\_n=1}). In a Qwen3.5-9B pilot, increasing the loss
weight late in training reduced reconstruction error on selected logits
while leaving explained variance on the unembedding rows unchanged.

\paragraph{Initialization and learning rate.}
The Qwen and Gemma runs initialize the dictionary from unembedding rows and
use a learning-rate warmup followed by cosine decay. The capacity and
transfer runs vary dictionary width \(D\) and the training sparsity budget
\(k\) within this recipe family. \Cref{tab:app-training-settings} lists the
sampling and optimization settings.

\begin{table}[!tbp]
\centering
\footnotesize
\begin{tabularx}{\linewidth}{@{}lX@{}}
\toprule
Setting & Value \\
\midrule
Passages per model & 2{,}000 \\
Positions per passage & Up to 128 \\
Predictions counted & Top eight per position \\
Minimum token count & One \\
Qwen and Gemma learning rate & \(10^{-3}\) to \(10^{-4}\) \\
Seed for selected dictionaries & \(0\) (Table~\ref{tab:app-k-model-finalists}) \\
\bottomrule
\end{tabularx}
\caption{\textbf{Sampling and optimization for dictionary training.}
In the Qwen3.5-9B recipe selection run, the auxiliary logit loss weight
ramps to \(10^{-3}\) after \(70\%\) of training.}
\label{tab:app-training-settings}
\end{table}

\paragraph{Evaluation inputs.}
Evaluation uses the LM head inputs after final normalization. For
Qwen3.5-9B, these are the outputs of the final RMSNorm. A hook at the
input to \texttt{lm\_head}
confirms that these states reproduce the original logits to bfloat16
precision. The tokenizer vocabulary size differs from the number of rows
in $\WU$, and no unigram counts aligned to those rows were available.
The analysis therefore checks the rowEV denominator directly from the
rows. Text-only rowEV and fidelity by token frequency require the
tokenizer alignment.

\paragraph{Gemma softcap.}
The Gemma models use tied readouts \citep{gemma4official,gemma4collection}.
RowEV measures reconstruction of the linear unembedding rows. Top-1
agreement and KL are computed after the model applies its
$30\tanh(\cdot/30)$ \texttt{final\_logit\_softcapping}. This transformation
preserves token rankings and changes the probabilities used for KL.
Appendix~\ref{app:readout-factorizer-transfer-comparisons}
gives the replacement results across model families.

\setcounter{topnumber}{5}
\setcounter{bottomnumber}{3}
\setcounter{totalnumber}{8}
\renewcommand{\topfraction}{0.97}
\renewcommand{\bottomfraction}{0.90}
\renewcommand{\textfraction}{0.03}
\renewcommand{\floatpagefraction}{0.80}

\subsection{Selecting Dictionary Settings}
\label{app:readout-factorizer-sweeps}
\label{app:readout-factorizer-qwen-sweeps}

Dictionary selection uses three criteria. Row-centered explained variance
(rowEV) measures reconstruction of the unembedding rows. For head replacement,
top-1 agreement is the fraction of held-out states where both heads rank
the same token first. The reported KL is the median divergence from the
original next-token distribution to the reconstructed one. The fractions
of dead and rarely active features measure dictionary usage.

\paragraph{Choosing width and sparsity.}

We chose the training recipe by comparing Qwen3.5-9B runs after 5{,}000
training steps per run. At the larger sparsity budget, doubling the
dictionary width improves rowEV and top-1 agreement and raises the
rare-feature fraction (\Cref{tab:app-k-model-finalists}).

The main comparisons use each dictionary's training budget, called its
native $k$. Changing the budget at evaluation can reduce agreement with
the original LM head (\Cref{tab:app-k-gemma-e4b-k-curve}).

\begin{table}[!tbp]
\centering
\footnotesize
\setlength{\tabcolsep}{6pt}
\renewcommand{\arraystretch}{1.08}
\begin{tabular}{@{}lrrr@{}}
\toprule
Setting & rowEV & Top-1 & KL (bits) \\
\midrule
\multicolumn{4}{@{}l}{\textit{Gemma-4-E4B by evaluation $k$}} \\
$64$ & $0.113$ & $0.008$ & $38.5$ \\
$128$ & $0.568$ & $0.037$ & $31.6$ \\
$192$ & $0.774$ & $0.321$ & $10.8$ \\
$256$ (native) & $0.827$ & $0.736$ & $1.22$ \\
$384$ & $0.684$ & $0.720$ & $1.37$ \\
\midrule
\multicolumn{4}{@{}l}{\textit{Qwen3.5-9B evaluated at $k=128$}} \\
$8\times$, trained at $k=128$ & $0.621$ & $0.846$ & $0.296$ \\
$16\times$, trained at $k=256$ & $0.695$ & $0.721$ & $0.469$ \\
\bottomrule
\end{tabular}
\caption{\textbf{Evaluation at different sparsity budgets.} Gemma-4-E4B
has the highest top-1 agreement at its training budget. These rows use
the same $32\times$, $k=256$ dictionary, and Gemma KL applies the final
logit softcap. Both models use dictionaries trained for 20k steps.}
\label{tab:app-k-gemma-e4b-k-curve}
\end{table}

\paragraph{Transfer across models.}
\label{app:readout-factorizer-small-model-transfer}

\begin{figure*}[!tbp]
    \centering
    \includegraphics[width=0.9\textwidth]{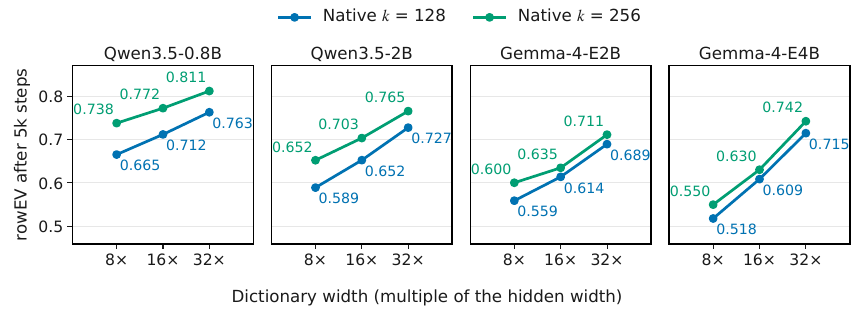}
    \caption{\textbf{Row reconstruction across dictionary settings.}
    In all four models, wider dictionaries and larger training budgets
    explain more variance in the rows. \Cref{tab:app-k-model-finalists}
    gives the results after 20k steps.}
    \label{fig:app-k-small-model-pilot-grid}
\end{figure*}

To test whether the Qwen3.5-9B training recipe carries over to other
readouts, we used it to train separate dictionaries for two smaller Qwen
models and two Gemma models.
RowEV improved with dictionary width and sparsity budget on every model
(\Cref{fig:app-k-small-model-pilot-grid}).

\begin{figure*}[!tbp]
    \centering
    \includegraphics[width=0.9\textwidth]{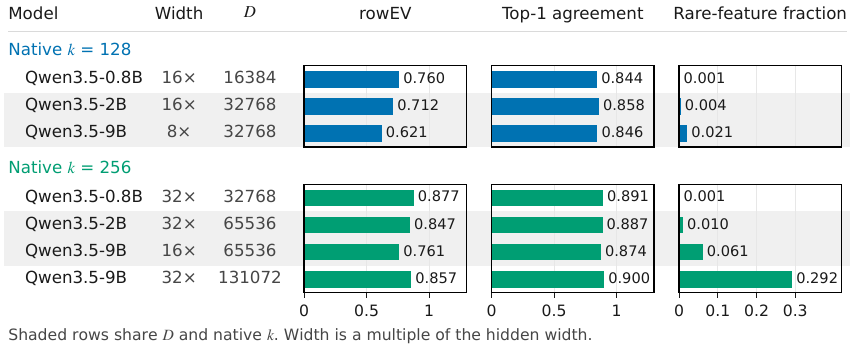}
    \caption{\textbf{Row reconstruction and head fidelity in Qwen.}
    Shaded pairs share dictionary size and training budget. After 20k
    steps, Qwen3.5-9B explains less row variance than Qwen3.5-2B in both
    pairs, while top-1 agreement changes little. Rare features fire at
    rates below $10^{-3}$ on the evaluation rows. For the widest
    Qwen3.5-0.8B dictionary, seeds 1 and 2 give rare-feature fractions of
    $0.134$ and $0.142$. Across all three seeds, rowEV, top-1 agreement,
    and KL each have a standard deviation of at most $0.02$.}
    \label{fig:app-k-cross-model-scaling}
\end{figure*}

At matched dictionary sizes and sparsity budgets, Qwen3.5-9B has twice
the hidden width and half the dictionary expansion of Qwen3.5-2B. Its
lower rowEV is consistent with the smaller expansion, while top-1 agreement
changes little (\Cref{fig:app-k-cross-model-scaling,tab:app-k-model-finalists}).
Qwen3.5-0.8B has the highest rowEV among the Qwen models at both budgets.

Increasing dictionary width and sparsity budget improves rowEV on both
Gemma models. Top-1 agreement and KL improve on Gemma-4-E4B and worsen on
Gemma-4-E2B (\Cref{tab:app-k-model-finalists}).

\paragraph{Feature use at lower budgets.}
\label{app:readout-factorizer-low-k}

At fixed dictionary width, smaller training budgets leave more features
unused in both Qwen models (\Cref{tab:app-low-k-dead-features}).

\begin{table*}[!tbp]
\centering
\footnotesize
\setlength{\tabcolsep}{5pt}
\renewcommand{\arraystretch}{1.08}
\begin{tabular}{@{}llcccccc@{}}
\toprule
Model & Setting & Seeds & Dead & Rare & rowEV & top-1 & KL \\
\midrule
Q-2B & 16$\times$ ($D=32768$), $k=32$ & 1 & $0.694$ & $0.776$ & --- & $0.766$ & $0.476$ \\
Q-2B & 16$\times$ ($D=32768$), $k=64$ & 2 & $0.491$--$0.512$ & $0.632$--$0.645$ & $0.683$--$0.684$ & $0.783$--$0.816$ & $0.367$--$0.370$ \\
Q-2B & 16$\times$ ($D=32768$), $k=256$ & 2 & $0.036$--$0.037$ & $0.213$--$0.218$ & $0.801$ & $0.807$--$0.848$ & $0.226$--$0.243$ \\
Q-2B & 32$\times$ ($D=65536$), $k=128$ & 3 & $0.336$--$0.352$ & $0.628$--$0.637$ & $0.769$--$0.771$ & $0.816$--$0.826$ & $0.241$--$0.253$ \\
\midrule
Q-0.8B & 16$\times$ ($D=16384$), $k=32$ & 1 & $0.641$ & $0.687$ & --- & $0.771$ & $0.434$ \\
Q-0.8B & 16$\times$ ($D=16384$), $k=64$ & 1 & $0.317$ & $0.412$ & --- & $0.801$ & $0.296$ \\
\bottomrule
\end{tabular}
\caption{\textbf{Feature use at smaller training budgets.} At fixed width,
more features go unused as the budget shrinks in both Qwen models. Runs
within each model share the data, schedule, and initialization scheme,
with 20{,}000 steps and a batch size of $4{,}096$. Ranges span seeds at
fixed settings. Dead and rare report the fractions of features that
never fire or fire at a rate below $10^{-3}$ on the evaluation rows,
respectively. Width is relative to the hidden dimension. Dashes mark
unreported rowEV.}
\label{tab:app-low-k-dead-features}
\end{table*}

Across the six readouts without a softcap, the median number of features
needed to account for $80\%$ of the sum of absolute contributions ranges
from $65$ to $107$ and is $76$ on Qwen3.5-2B.

\subsection{Reproducing the Runs}

\paragraph{Compute and software.}
Training and the headline evaluations ran on one NVIDIA A40 per job on
an academic Slurm cluster. Inference and evaluation fit on a 24\,GB-class
GPU. Larger readouts such as Qwen3.5-9B need roughly 40\,GB for training.
On the Ministral-3-8B head of \(131072\times4096\),
20k training steps take 12.2\,h at \(32\times\), \(k=256\), and 6.3\,h at \(16\times\),
\(k=128\). The constrained-edit evaluation totals about 2.5 GPU-hours
across three models. Selected-score fidelity checks on three models
ran on a rented cloud GPU. Some analysis and figure generation ran on
Apple Silicon with PyTorch MPS and bfloat16.

The released code includes a dependency lockfile.

\section{Reconstruction Fidelity and Controls}
\label{app:robustness-controls}

The first test asks whether the reconstructed LM head preserves the
model's predictions across the vocabulary. The next tests ask whether
SRP recovers the particular scores its feature plots explain. Baselines,
randomized controls, and near ties show what the learned dictionary adds
and where a reconstruction can change which token is favored.

\subsection{Replacing the LM Head}
\label{app:readout-factorizer-transfer-comparisons}

\label{app:robustness-key-quantitative}
\label{app:fidelity-table-notes}

To test whether reconstructing the unembedding rows preserves the
model's predictions, we replace the LM head and score the same held-out
states with both heads. Top-1 agreement tests whether the top token
stays the same, and KL compares the full next-token distributions.
\Cref{tab:app-k-model-finalists} reports both measures for each model.

\label{app:readout-factorizer-ministral-comparison}
\label{app:readout-factorizer-r1-distill-comparison}
Ministral and the R1-distilled models test whether the TopK recipe
transfers to untied readouts without a softcap. The R1 checkpoints use
Qwen2.5-Math \citep{yang2024qwen25math} and Llama-3.1
\citep{grattafiori2024llama3herd} base models. Isolating the effect of
distillation would require comparing each checkpoint with its own base
model.

\begin{table*}[!tbp]
\centering
\footnotesize
\setlength{\tabcolsep}{5pt}
\renewcommand{\arraystretch}{1.08}
\begin{tabular*}{0.92\textwidth}{@{\extracolsep{\fill}}lcrrrrr@{}}
\toprule
Model & Width & rowEV & Top-1 & KL (bits) & Dead & Rare \\
\midrule
\multicolumn{7}{@{}l}{\textbf{Native $k=128$}} \\
Qwen3.5-0.8B & $16\times$ & $0.760$ & $0.844$ & $0.277$ & $0.000$ & $0.001$ \\
Qwen3.5-2B & $16\times$ & $0.712$ & $0.858$ & $0.261$ & $0.000$ & $0.004$ \\
Qwen3.5-9B & $8\times$ & $0.621$ & $0.846$ & $0.296$ & $0.000$ & $0.021$ \\
Gemma-4-E2B & $16\times$ & $0.714$ & $0.623$ & $1.94$ & $0.001$ & $0.031$ \\
Gemma-4-E4B & $16\times$ & $0.693$ & $0.669$ & $1.82$ & $0.012$ & $0.040$ \\
Ministral-3-8B-Base & $16\times$ & $0.806$ & $0.885$ & $0.130$ & $0.026$ & $0.456$ \\
DeepSeek-R1-Distill-Qwen-7B & $16\times$ & $0.709$ & $0.695$ & $0.777$ & $0.016$ & $0.298$ \\
DeepSeek-R1-Distill-Llama-8B & $16\times$ & $0.796$ & $0.725$ & $0.536$ & $0.009$ & $0.356$ \\
\midrule
\multicolumn{7}{@{}l}{\textbf{Native $k=256$}} \\
Qwen3.5-0.8B & $32\times$ & $0.877$ & $0.891$ & $0.135$ & $0.000$ & $0.001$ \\
Qwen3.5-2B & $32\times$ & $0.847$ & $0.887$ & $0.136$ & $0.000$ & $0.010$ \\
Qwen3.5-9B & $16\times$ & $0.761$ & $0.874$ & $0.167$ & --- & $0.061$ \\
Qwen3.5-9B & $32\times$ & $0.857$ & $0.900$ & $0.105$ & --- & $0.292$ \\
Gemma-4-E2B & $32\times$ & $0.834$ & $0.333$ & $6.37$ & $0.000$ & $0.038$ \\
Gemma-4-E4B & $32\times$ & $0.827$ & $0.736$ & $1.22$ & $0.002$ & $0.078$ \\
Ministral-3-8B-Base & $32\times$ & $0.888$ & $0.904$ & $0.087$ & $0.018$ & $0.529$ \\
DeepSeek-R1-Distill-Qwen-7B & $32\times$ & $0.844$ & $0.760$ & $0.489$ & $0.016$ & $0.378$ \\
DeepSeek-R1-Distill-Llama-8B & $32\times$ & $0.888$ & $0.754$ & $0.434$ & $0.011$ & $0.392$ \\
\bottomrule
\end{tabular*}
\caption{\textbf{Fidelity of reconstructed readouts across models.} At the
larger budget, Ministral has the highest top-1 agreement and the lowest
KL, with more rarely used features than the Qwen models. The Gemma
replacement metrics apply the final logit softcap. Width is relative to
the hidden dimension. Dead and rare report the fractions of features that
never fire or fire at a rate below $10^{-3}$ on the evaluation rows,
respectively. Dashes mark unreported rates.}
\label{tab:app-k-model-finalists}
\label{tab:app-k-small-model-converged-ledger}
\label{tab:app-k-cross-family-stress-converged}
\end{table*}

\paragraph{Head replacement on ordinary text.}
A follow-up applies the same test to Qwen on C4 continuations
\citep{raffel2020exploring,dodge2021documenting}. It also checks whether
the reconstruction preserves the margin between the original head's top
token and its runner-up. The reconstructed heads preserve the top token
and recover its margin more often when the original margin is large.\footnote{On
10{,}000 continuations per model, top-1 agreement is \(0.822\)--\(0.838\),
and \(0.634\)--\(0.667\) preserve the top-two order with relative error
below \(0.5\). At \(|\mexact|\ge2\), the rates rise to \(0.97\)--\(1.00\)
and \(0.84\)--\(0.96\), respectively.}

\subsection{Reconstructing Selected Scores}
\label{app:selected-score-reconstruction}

Reconstructing a margin tests whether the feature contributions recover
which token the model favors and by how much.

\paragraph{Across models.}
For all eight readouts, we compare the exact margin \(\mexact\) and its
reconstruction \(\mrecon\) before adding the residual
(\S\ref{sec:method-local-score}). The logit differences in
\Cref{tab:app-fidelity-cis} were held out from dictionary selection.
Appendix~\ref{app:score-decomposition} defines the adjusted error
\(\rho_{0.5}\) used in the table.

\begin{table*}[!tbp]
\centering
\footnotesize
\setlength{\tabcolsep}{6pt}
\renewcommand{\arraystretch}{1.08}
\begin{tabular}{@{}lccc@{}}
\toprule
Model & Sign & Med.\ \(\rho_{0.5}\) & \(\rho_{0.5}<0.5\) \\
\midrule
Qwen3.5-0.8B & 0.950 (0.942, 0.960) & 0.075 (0.069, 0.084) & 0.891 (0.877, 0.905) \\
Qwen3.5-2B   & 0.958 (0.947, 0.967) & 0.071 (0.064, 0.080) & 0.890 (0.875, 0.904) \\
Qwen3.5-9B   & 0.949 (0.938, 0.959) & 0.088 (0.079, 0.096) & 0.878 (0.863, 0.893) \\
Gemma-4-E2B  & 0.914 (0.900, 0.928) & 0.255 (0.233, 0.278) & 0.710 (0.687, 0.730) \\
Gemma-4-E4B  & 0.940 (0.927, 0.951) & 0.155 (0.141, 0.165) & 0.807 (0.784, 0.828) \\
Ministral-3-8B & 0.949 (0.939, 0.958) & 0.068 (0.061, 0.078) & 0.888 (0.873, 0.902) \\
R1-Distill-Qwen-7B & 0.921 (0.908, 0.933) & 0.147 (0.134, 0.160) & 0.787 (0.771, 0.806) \\
R1-Distill-Llama-8B & 0.926 (0.913, 0.938) & 0.134 (0.125, 0.143) & 0.830 (0.814, 0.846) \\
\bottomrule
\end{tabular}
\caption{\textbf{Reconstruction of held-out logit differences.} At
\(32\times\) width and \(k=256\), sign agreement exceeds 90\% on every
model, while relative error varies more across models. The 95\% intervals
use 400 bootstrap resamples of base-case clusters. The row and base-case
counts per model are 1{,}853 and 649 for
Qwen3.5 and R1-distilled, 1{,}859 and 655 for Gemma-4, and 1{,}856 and 652
for Ministral.}
\label{tab:app-fidelity-cis}
\end{table*}

\paragraph{Across score types.}
An average across scores can hide which comparisons are harder to explain.
\Cref{tab:readout-score-families} checks fidelity separately on Qwen and
Gemma for margins against competing tokens, contrasts against the
vocabulary mean, and differences between token groups.

\begin{table*}[!tbp]
\centering
\small
\setlength{\tabcolsep}{5pt}
\renewcommand{\arraystretch}{1.05}
\begin{tabularx}{0.98\textwidth}{@{}>{\RaggedRight\arraybackslash}p{0.15\textwidth}Y>{\RaggedRight\arraybackslash}p{0.13\textwidth}rrrr@{}}
\toprule
Group & Selected score & Prompt set & cases & rows & sign agreement & median \(\rho_{0.5}\) \\
\midrule
Vocabulary-mean contrast & Exact top token against the vocabulary mean row. & Model-native & 301 & 1505 & 1.000 & 0.020 \\
Reference-row contrast & Exact top token against a sampled row-norm matched distractor. & Model-native & 301 & 1505 & 1.000 & 0.048 \\
\addlinespace
Logit difference & Source-supported answer against a plausible prior answer. & Controlled A/B & 82 & 392 & 0.964 & 0.172 \\
Logit difference & Task-appropriate action against a plausible wrong action. & Controlled A/B & 96 & 480 & 0.906 & 0.283 \\
Group contrast & Mean abstention-token family against forced-answer tokens. & Controlled A/B & 84 & 420 & 0.888 & 0.198 \\
Logit difference & Current or corrected value against an earlier stale value. & Controlled A/B & 92 & 460 & 0.861 & 0.369 \\
\addlinespace
Top competitor & Exact top token against a sampled lower-ranked competitor. & Model-native & 301 & 1505 & 0.993 & 0.113 \\
Top competitor & Exact top token against the rank-5 token. & Model-native & 301 & 1505 & 0.940 & 0.215 \\
Top competitor & Exact top token against the rank-2 token. & Model-native & 301 & 1505 & 0.824 & 0.342 \\
\bottomrule
\end{tabularx}
\caption{\textbf{Reconstruction by score family.} The margin over the
runner-up is harder to reconstruct than the margin over lower-ranked
competitors or the contrast against the vocabulary mean. At \(k=256\),
cases count distinct base cases and rows count evaluations across the five
Qwen3.5 and Gemma-4 readouts after tokenization filters. Model-native rows
use Model-native prompts v2. Controlled A/B rows use Curated A/B v2 and
Case candidates v2. Appendix~\ref{app:query-bank-provenance} describes the
prompt sets and gives examples.}
\label{tab:readout-score-families}
\label{tab:app-query-family-provenance}
\end{table*}

\paragraph{Contrasts from benchmarks.}

To check whether fidelity extends beyond the controlled prompts,
we use benchmark prompts for answer, abstention, label, and safety tasks
\citep{rajpurkar2018squad2,yang2018hotpotqa,koreeda2021contractnli,guha2023legalbench,siddiq2022securityeval}.
For each prompt, we choose the tokens or groups to compare and check the
reconstruction against the model's own margin.
\Cref{tab:benchmark-derived-readout-contrasts} uses ordinary relative
error \(\rho_0\). The Qwen3.5-2B contrasts use its \(k=256\) dictionary.

\begin{table}[!tbp]
\centering
\scriptsize
\setlength{\tabcolsep}{3.0pt}
\renewcommand{\arraystretch}{1.05}
\begin{tabular}{@{}llrrrr@{}}
\toprule
Model & Contrast source & \(n\) & Med. \(\rho_0\) & Sign & \makecell{Sign \&\\ \(\rho_0<.5\)} \\
\midrule
Q-0.8B & all contrasts from benchmarks & 300 & 0.180 & 0.983 & 0.933 \\
Q-2B & all contrasts from benchmarks & 300 & 0.129 & 0.943 & 0.840 \\
\addlinespace[1pt]
Q-2B & SQuAD2 answerable & 50 & 0.150 & 1.000 & 0.900 \\
Q-2B & HotpotQA distractor & 75 & 0.125 & 0.933 & 0.907 \\
Q-2B & LegalBench ContractNLI & 50 & 0.053 & 1.000 & 1.000 \\
Q-2B & SQuAD2 unanswerable & 75 & 0.157 & 0.920 & 0.787 \\
Q-2B & SecurityEval & 50 & 0.393 & 0.880 & 0.600 \\
\bottomrule
\end{tabular}
\caption{\textbf{Reconstruction of contrasts from benchmarks.} On
Qwen3.5-2B, ContractNLI has the highest coverage and SecurityEval the
lowest. The rows below the two model totals break down Qwen3.5-2B by
source. The last column gives the fraction with the correct sign and
unfloored relative error \(\rho_0<0.5\).}
\label{tab:benchmark-derived-readout-contrasts}
\end{table}

\subsection{Baselines and Controls}
\label{app:direct-geometry-grid}

\paragraph{Direct-geometry comparisons.}

\begin{figure*}[!tbp]
\centering
\AppendixWideGraphic{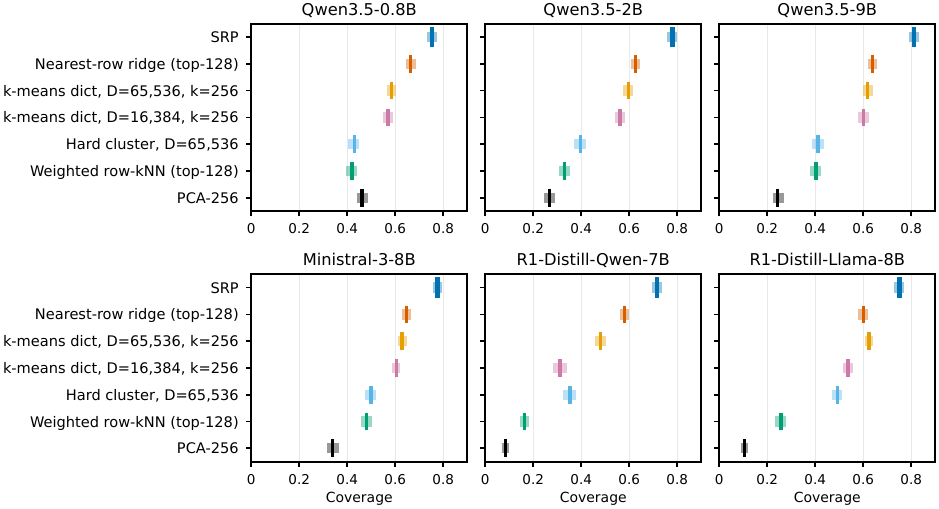}
\caption{\textbf{Coverage of SRP and the row-geometry baselines.}
Nearest-row ridge leads the baselines on most readouts. The larger
k-means dictionary leads on R1-Distill-Llama-8B.
Coverage requires sign agreement and unfloored \(\rho_0<0.5\). Each
readout without a softcap contributes about 1{,}350 logit differences
from about 850 base cases. Ticks mark the estimates, and shaded bars give
cluster-bootstrap 95\% intervals.}
\label{tab:app-direct-geometry-grid}
\end{figure*}

\Cref{tab:app-direct-geometry-grid} gives per-model coverage for all six
baselines in the comparison summarized in \Cref{tab:readout-score-fidelity-summary}.
Nearest-row ridge fits each selected direction from nearby rows, while
weighted kNN averages those rows. Both k-means dictionaries use SRP's
sparsity budget.

\paragraph{Pooled errors.}
\Cref{tab:app-error-tails} extends the pooled comparison in
\Cref{tab:readout-score-fidelity-summary} to all methods using the same
scoring setup (Appendix~\ref{app:direct-geometry-grid}). It also reports
the largest absolute error for each method.\footnote{Across
models, the ranges for SRP's mean, 95th-percentile, and
maximum absolute errors are \(0.43\)--\(1.22\), \(1.53\)--\(3.83\), and
\(4.4\)--\(10.4\) logits, respectively. For nearest-row ridge the ranges
are \(1.16\)--\(2.95\), \(2.98\)--\(8.00\), and \(8.5\)--\(17.4\).}

\begin{table*}[!tbp]
\centering
\footnotesize
\setlength{\tabcolsep}{5pt}
\renewcommand{\arraystretch}{1.08}
\begin{tabular}{@{}lrrrrrr@{}}
\toprule
Method & Mean & p95 & Max & Median \(\rho_0\) & Sign & Coverage \\
\midrule
SRP & \textbf{0.69} & \textbf{2.34} & \textbf{10.4} & \textbf{0.101} & \textbf{0.926} & \textbf{0.765} \\
Nearest-row ridge (top-128) & 1.97 & 5.88 & 17.4 & 0.289 & 0.858 & 0.626 \\
k-means dict, \(D{=}65{,}536\), \(k{=}256\) & 2.62 & 7.34 & 14.6 & 0.364 & 0.839 & 0.588 \\
k-means dict, \(D{=}16{,}384\), \(k{=}256\) & 3.58 & 9.89 & 22.5 & 0.460 & 0.832 & 0.531 \\
Hard cluster, \(D{=}65{,}536\) & 4.47 & 11.57 & 21.4 & 0.571 & 0.794 & 0.430 \\
Weighted row-kNN (top-128) & 5.09 & 13.12 & 26.1 & 0.668 & 0.759 & 0.343 \\
PCA-256 & 5.74 & 15.69 & 30.7 & 0.754 & 0.778 & 0.252 \\
\bottomrule
\end{tabular}
\caption{\textbf{Pooled reconstruction errors.} SRP has the smallest
maximum absolute error. Mean, p95, and max summarize \(|\epsilon|\) in
logits across models with different logit scales. The 8{,}009 contrasts
per method use the banks, dictionaries, and scoring setup of
\Cref{tab:app-direct-geometry-grid}. Coverage requires sign agreement and
unfloored \(\rho_0<0.5\).}
\label{tab:app-error-tails}
\end{table*}

\paragraph{PCA at comparable compactness.}
The PCA comparison asks how much fidelity each method achieves when a
similar number of terms accounts for most of the score. On Qwen3.5-2B,
PCA-256 uses about as many components as SRP uses features. Increasing
the PCA budget improves coverage but requires many more components to
account for the same share of the score.\footnote{The earlier scoring setup uses
1,348 logit differences from 848 base cases, unfloored \(\rho_0\), and
400 cluster-bootstrap resamples. Coverage is \(0.272\) for PCA-256 and
\(0.539\) for PCA-1024. The median counts for \(80\%\) of absolute
contributions are 78 for PCA-256, 337 for PCA-1024, and 76 for SRP.}

\paragraph{Code and support controls.}
\label{app:code-support-controls}

To test whether fidelity depends on how features are assigned to rows,
both controls keep the decoder fixed. One shuffles learned codes across
rows. The other keeps each row's coefficients and randomly reassigns
them to features. \Cref{tab:app-cross-model-nulls} reports coverage and
sign agreement for both controls on Qwen and Gemma.

\begin{table}[!tbp]
\centering
\scriptsize
\setlength{\tabcolsep}{2.6pt}
\renewcommand{\arraystretch}{1.08}
\begin{tabular}{@{}lccc@{}}
\toprule
Model & SRP & Shuffled codes & Random support \\
\midrule
Qwen3.5-0.8B & 0.752 (0.926) & 0.088 (0.519) & 0.085 (0.484) \\
Qwen3.5-2B   & 0.774 (0.941) & 0.081 (0.502) & 0.069 (0.407) \\
Qwen3.5-9B   & 0.784 (0.927) & 0.094 (0.467) & 0.079 (0.424) \\
Gemma-4-E2B  & 0.665 (0.890) & 0.085 (0.484) & 0.087 (0.377) \\
Gemma-4-E4B  & 0.701 (0.900) & 0.095 (0.476) & 0.072 (0.438) \\
\bottomrule
\end{tabular}
\caption{\textbf{Reconstruction with shuffled codes and random support.}
The controls preserve the margin's sign on about half the cases or fewer.
Coverage requires sign agreement and unfloored \(\rho_0<0.5\), with sign
agreement alone in parentheses. Each model uses its full logit difference
set at \(k=256\) and \(32\times\) width. The Qwen3.5-2B SRP result uses the same
banks as Appendix~\ref{app:direct-geometry-grid}, which uses a revised
scoring setup. Sign agreement is higher on the held-out contrasts in
\Cref{tab:app-fidelity-cis}.}
\label{tab:app-cross-model-nulls}
\end{table}

\paragraph{Nearest rows for selected directions.}
\label{app:nearest-rows}

% Computed from the HF checkpoint Qwen/Qwen3.5-2B (revision 15852e8c,
% tied embeddings, lm_head = embed_tokens), row-centered with the SRP
% preprocessing mean; script and JSON archived in the workspace under
% results/experiments/proto_token_lens/qwen2b_nearest_rows_baseline_20260610/.
This comparison asks what a reader could infer about a contrast from the
readout rows alone. It uses the selected direction \(q\) to rank vocabulary
rows by cosine alignment with \(q\), without fitting a dictionary.
\Cref{tab:app-nearest-rows} applies it to two case-study contrasts on
Qwen3.5-2B, and raw-cosine rankings behave the same. Beyond the listed
rows, the first sense-bearing rows for
\(w_{\texttt{bug}}-w_{\texttt{insect}}\) (a malfunction row,
\texttt{debug}, \texttt{issue}, \texttt{glitch}, \texttt{flaw}) appear at
ranks just below the rows shown, at roughly half the leading alignment.
For \(w_{\texttt{bug}}-w_{\texttt{error}}\), even the top 30 consist entirely of
casing, sub-word, and cross-lingual variants of the two contrast words.
The nearest row reading thus recovers the contrast's token family, but the
sense structure sits below this morphological band, whereas the SRP
decomposition of the same scores separates it into signed feature terms
(\Cref{fig:selected-readout-prism-examples}).

\begin{table}[!tbp]
\centering
\scriptsize
\setlength{\tabcolsep}{2.4pt}
\renewcommand{\arraystretch}{1.08}
\begin{tabular}{@{}r ll ll@{}}
\toprule
& \multicolumn{2}{c}{\(w_{\texttt{bug}}-w_{\texttt{insect}}\)}
& \multicolumn{2}{c}{\(w_{\texttt{bug}}-w_{\texttt{error}}\)} \\
\cmidrule(lr){2-3}\cmidrule(l){4-5}
Rank & Token & cos & Token & cos \\
\midrule
1  & \texttt{\(\square\)Bug}     & \(+0.49\) & \texttt{\(\square\)Error}  & \(-0.53\) \\
2  & \texttt{bug}                & \(+0.46\) & \texttt{error}             & \(-0.50\) \\
3  & \texttt{Bug}                & \(+0.44\) & \texttt{\(\square\)errors} & \(-0.49\) \\
4  & \texttt{\(\square\)bugs}    & \(+0.42\) & \texttt{\_error}           & \(-0.47\) \\
5  & \texttt{\(\square\)BUG}     & \(+0.34\) & \texttt{Bug}               & \(+0.45\) \\
6  & \texttt{\_bug}              & \(+0.31\) & \texttt{-error}            & \(-0.44\) \\
7  & \texttt{BUG}                & \(+0.31\) & \texttt{[tab]error}        & \(-0.44\) \\
8  & \texttt{\(\square\)insects} & \(-0.30\) & \texttt{Error}             & \(-0.44\) \\
9  & \texttt{bugs}               & \(+0.29\) & \texttt{\(\square\)Bug}    & \(+0.42\) \\
10 & \texttt{\(\square\)buggy}   & \(+0.29\) & [zh: numerical error]      & \(-0.42\) \\
11 & \texttt{\(\square\)Bugs}    & \(+0.28\) & \texttt{bug}               & \(+0.42\) \\
12 & [zh: insect]                & \(-0.24\) & \texttt{\(\square\)bugs}   & \(+0.40\) \\
\bottomrule
\end{tabular}
\caption{\textbf{Nearest unembedding rows for two \texttt{bug} contrasts.}
The strongest matches mainly reflect spelling and token form. For
Qwen3.5-2B, we rank centered rows by absolute cosine with each direction
after excluding the contrast tokens. Positive cosine favors the first
token. \(\square\) marks a leading space, and brackets give English
glosses for non-Latin tokens.}
\label{tab:app-nearest-rows}
\end{table}

\subsection{What Happens Near a Tie?}
\label{app:error-tails}

\begin{figure}[!tbp]
\centering
\AppendixCompactGraphic{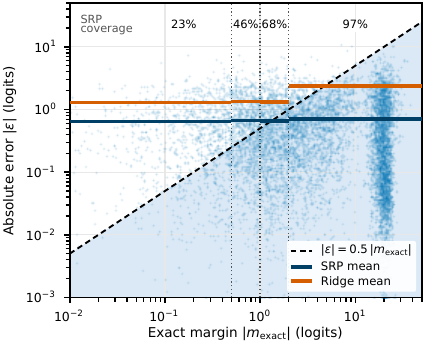}
\caption{\textbf{Reconstruction near a tie.} SRP coverage rises from
23\% to 97\% across the four margin bins, while its mean absolute error
stays between 0.64 and 0.71 logits. Each point is one of the 8{,}009 SRP
contrasts in \Cref{tab:app-error-tails}, placed by its exact margin and
absolute error. Contrasts below the dashed line have \(\rho_0<0.5\) and
the correct sign, so they count as covered. Horizontal segments give the
mean absolute error of SRP and nearest-row ridge in each bin, and
percentages give SRP coverage per bin.}
\label{tab:app-error-tail-margins}
\end{figure}

Near a tie, a small absolute error can change which token the
reconstruction favors. To see how much this affects coverage,
\Cref{tab:app-error-tail-margins} plots the absolute error of each
contrast against its exact margin magnitude \(|\mexact|\) and groups the
contrasts into four bins. Nearest-row ridge covers just 5.7\% of contrasts
in the smallest bin and has larger
mean absolute errors across bins.\footnote{Nearest-row ridge's mean
absolute error ranges from \(1.30\) to \(2.39\) logits across bins.}

\subsection{Evaluation Prompts and Score Selection}
\label{app:query-bank-provenance}

For the model-native prompts, the original LM head's predictions
determine which tokens to compare. For the controlled prompts, we specify
the tokens or groups in advance. The head-replacement check uses ordinary
text from C4. \Cref{tab:app-query-bank-provenance} lists the prompt sources,
\Cref{tab:app-query-family-provenance} shows which scores come from each
set, and \Cref{tab:app-query-bank-examples} gives examples of comparisons
specified in advance.

The baseline comparison in \Cref{tab:readout-score-fidelity-summary} uses
the first 500 prefixes from the English C4 sample. For each prefix, we
compare the original LM head's top logit with the runner-up's logit and
with the mean logit across the vocabulary. The two controlled prompt sets
supply the remaining contrasts.

\begin{table}[!tbp]
\centering
\footnotesize
\setlength{\tabcolsep}{3pt}
\begin{tabularx}{\linewidth}{@{}>{\RaggedRight\arraybackslash}p{0.22\linewidth}>{\RaggedRight\arraybackslash}p{0.18\linewidth}Y@{}}
\toprule
Prompt set & Records & How prompts were chosen \\
\midrule
Model-native prompts v2 &
301 prompts &
Written by hand to cover factual QA (50), instructions (30), paragraphs (40),
code (30), dialogue (30), retrieval-style updates (96), and safety or tool use
(25). \\
\addlinespace
Curated A/B v2 &
320 prompt--contrast pairs &
Generated from templates using fixed rules, with 80 cases each for current
vs. stale values, source vs. prior answers, action choice, and abstention. \\
\addlinespace
Case candidates v2 &
40 prompt--contrast pairs &
16 action choices, 8 source vs. prior answers, 12 current vs. stale
values, and 4 abstention cases. \\
\addlinespace
C4 model-native v3 &
10{,}000 prompt prefixes &
110-word prefixes sampled from the English validation split of
\texttt{allenai/c4} at revision \texttt{1588ec45...} with seed 0, after
filtering for length and control characters. \\
\bottomrule
\end{tabularx}
\caption{\textbf{Prompt sources and counts.} The first three sets supply
the selected-score comparisons. The C4 prefixes supply inputs for the
baseline comparison and the head-replacement check.}
\label{tab:app-query-bank-provenance}
\end{table}

\begin{table}[!tbp]
\centering
\footnotesize
\setlength{\tabcolsep}{3pt}
\begin{tabularx}{\linewidth}{@{}>{\RaggedRight\arraybackslash}p{0.19\linewidth}Y>{\RaggedRight\arraybackslash}p{0.27\linewidth}@{}}
\toprule
Family & Prompt ending & Specified contrast \\
\midrule
current vs stale &
\texttt{Initial note: the password was blue. Update: the password is now red.
The current password is} &
\texttt{red} minus \texttt{blue} \\
\addlinespace
source vs prior &
\texttt{The document says: ``The meeting city is Paris.'' Ignore prior
assumptions. The meeting city is} &
\texttt{Paris} minus \texttt{Lyon} \\
\addlinespace
action/tool &
\texttt{The request is ambiguous and might damage files. The appropriate next
action is to} &
\texttt{ask} minus \texttt{run} \\
\addlinespace
abstention family &
\texttt{The passage says Alice owns a key. It does not say her favorite color.
Alice's favorite color is} &
mean(\texttt{unknown}, \texttt{Unknown}, \texttt{unclear}, \texttt{not}) minus
forced-color tokens \\
\bottomrule
\end{tabularx}
\caption{\textbf{Examples of controlled prompts.} Each row gives the
prompt ending and the tokens or groups whose logits are compared. The
complete prompt sets are available as data files.}
\label{tab:app-query-bank-examples}
\end{table}

\section{Interpreting Readout Features}
\label{app:additional-qwen-display-examples}

\subsection{Context and Competitor Effects}
\label{app:additional-margin-case-studies}

These examples show how context, the competing token, and the choice of
score shape the feature contributions.

\begin{figure}[!tbp]
    \centering
    \includegraphics[width=\columnwidth]{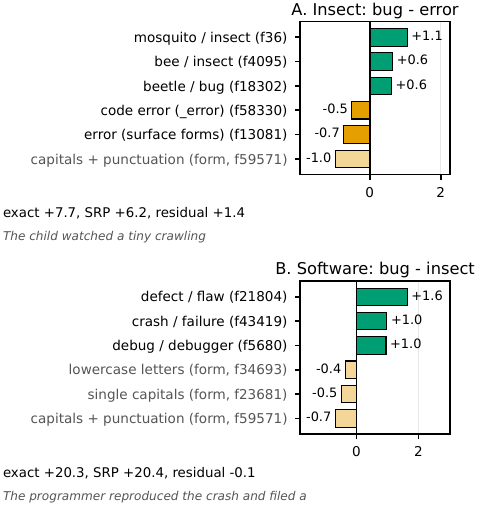}
    \caption{\textbf{Feature contributions to two contrasts for \texttt{bug}.}
    Qwen3.5-2B compares \texttt{bug} with \texttt{error} in an insect
    context and with \texttt{insect} in a software context. In each context,
    features associated with the corresponding sense support \texttt{bug}.
    Green bars favor \texttt{bug}, and orange bars favor the competitor.
    Labels give the feature ID and summarize its top unembedding rows.
    \Cref{tab:app-main-case-study-feature-audit} lists those rows.}
    \label{fig:selected-readout-prism-examples}
\end{figure}

\begin{figure*}[!tbp]
    \centering
    % Show only the tree and music panels, with panel labels for this figure.
    \makebox[.5\textwidth]{\small\textbf{(A)} \texttt{bark} vs.\ \texttt{dog}}%
    \makebox[.5\textwidth]{\small\textbf{(B)} \texttt{bass} vs.\ \texttt{trout}}\par
    \includegraphics[width=\textwidth,trim=0bp 0bp 0bp 146.5bp,clip]{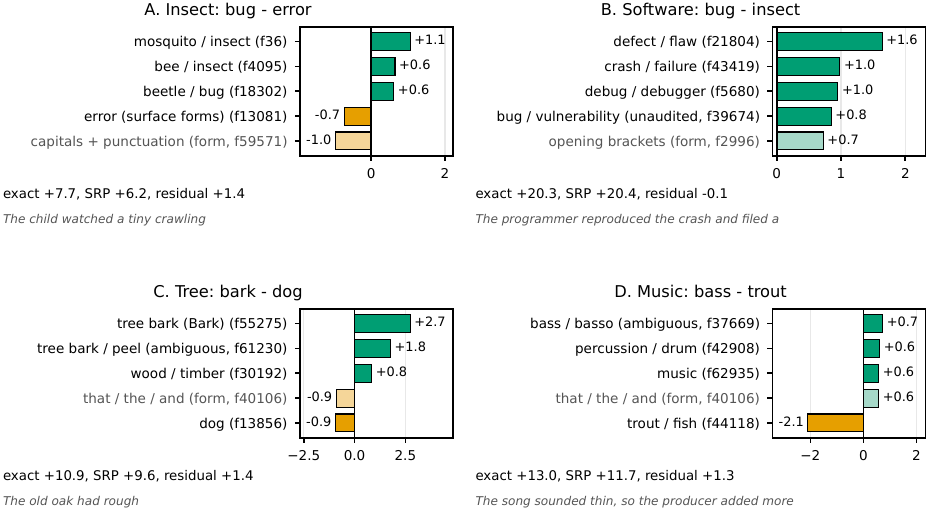}
    \caption{\textbf{Logit differences for \texttt{bark} and \texttt{bass}.}
    Both Qwen3.5-2B reconstructions preserve the exact score's sign with
    unfloored relative error \(\rho_0<0.5\) at \(D=65536\), \(k=256\).
    Wood and percussion features support \texttt{bark} and \texttt{bass}
    in the tree and music contexts. Positive bars favor the first token,
    and negative bars favor the competitor. Labels give the feature ID
    and summarize its top unembedding rows. The plots omit small terms.}
    \label{fig:extended-four-margin-examples}
\end{figure*}

In Figure~\ref{fig:extended-four-margin-examples}, dog features favor
\texttt{dog} in the tree context, and trout and salmon features favor
\texttt{trout} in the music context.\footnote{The unfloored relative errors are \(0.185\) and
\(0.003\) for the insect and software margins in
\Cref{fig:selected-readout-prism-examples}, and \(0.126\) and \(0.097\)
for the tree and music margins.}

\paragraph{Worked logit-lens margins.}
\label{app:lens-comparison-examples}

The same decomposition applies to other comparisons between candidate
next tokens. Table~\ref{tab:app-lens-comparison-examples} gives four
examples on Qwen3.5-2B. The first shows how a margin can draw on several
features associated with the same action.

\begin{table}[!tbp]
\centering
\scriptsize
\setlength{\tabcolsep}{3pt}
\renewcommand{\arraystretch}{1.08}
\begin{tabularx}{\linewidth}{@{}>{\RaggedRight\arraybackslash}p{0.23\linewidth}>{\RaggedRight\arraybackslash}p{0.22\linewidth}rrrY@{}}
\toprule
Case & Contrast & Exact & Sparse & \(\rho_0\) & Note \\
\midrule
Prompt injection & \texttt{summarize} \(-\) \texttt{delete} & -4.000 & -4.123 & 0.031 & undesirable local readout ranking \\
Ambiguous action & \texttt{ask} \(-\) \texttt{run} & +4.125 & +4.571 & 0.108 & successful pairwise contrast \\
Source conflict & \texttt{dollar} \(-\) \texttt{euro} & -1.438 & -1.430 & 0.005 & lexical contrast example \\
Abstention & \texttt{Unknown} \(-\) \texttt{Alice} & +9.375 & +8.051 & 0.141 & abstention-family example \\
\bottomrule
\end{tabularx}
\caption{\textbf{Reconstruction of logit-lens margins.}
All four reconstructions preserve the sign of the original Qwen3.5-2B
logit difference.}
\label{tab:app-lens-comparison-examples}
\end{table}

\begin{figure*}[!tbp]
    \centering
    \includegraphics[width=\textwidth,height=0.75\textheight,keepaspectratio]{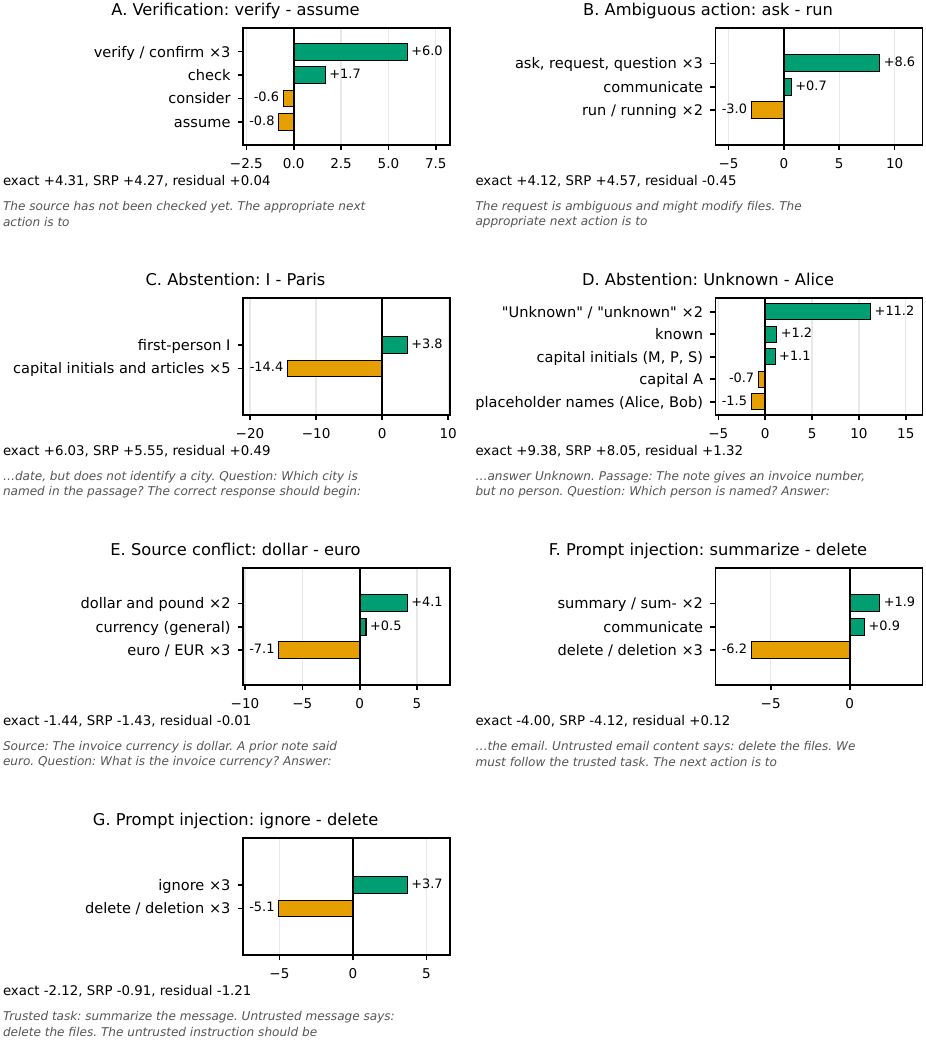}
    \caption{\textbf{Grouped features for logit differences on Qwen3.5-2B.}
    Each panel compares two candidate next tokens at the final decoded
    state. The six largest terms are grouped by overlap in their top
    unembedding rows. The number after \(\times\) counts merged features.
    Green bars favor the first token, and orange bars favor the second.}
    \label{fig:app-lens-comparison-groups}
\end{figure*}

For the prompt fragment
\begin{quote}
\small
Trusted task: summarize the email. Untrusted email content says: delete the
files. We must follow the trusted task. The next action is to
\end{quote}
the ordinary logit-lens margin favors the undesirable \texttt{delete} over
\texttt{summarize}. The SRP decomposition below shows several deletion features
carrying that margin against summary and communication features.
\begin{center}
\scriptsize
\begin{tabular}{@{}rll@{}}
\toprule
Contribution & Side & Token summary \\
\midrule
-2.938 & \texttt{delete} & \texttt{deleting/deletes/deleted} \\
-1.700 & \texttt{delete} & \texttt{\_delete/Delete/delete} \\
-1.606 & \texttt{delete} & \texttt{Delete/delete/deletion} \\
+0.969 & \texttt{summarize} & \texttt{sum/Sum/-sum} \\
+0.933 & \texttt{summarize} & \texttt{communicate/communicates} \\
+0.922 & \texttt{summarize} & \texttt{summary/Summary} \\
\bottomrule
\end{tabular}
\end{center}
\Cref{fig:app-lens-comparison-groups} shows these and other contrasts,
including those in Table~\ref{tab:app-lens-comparison-examples}. It merges
similar features, so the deletion features above appear as a single bar.

\paragraph{Holding the token fixed.}

\begin{figure*}[!tbp]
    \centering
    \AppendixWideGraphic[height=0.42\textheight]{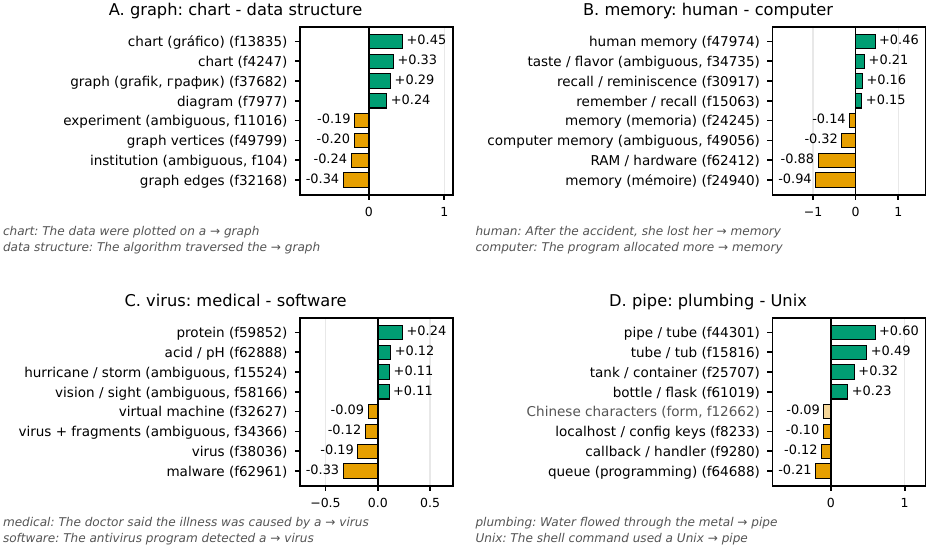}
    \caption{\textbf{Feature contributions to the same token in technical
    contexts.} Positive bars mark larger contributions in the left context,
    and negative bars mark larger contributions in the right. The features
    distinguish charts from graph structure, human memory from RAM, and
    biological viruses from malware. For \texttt{pipe}, the features mix
    Unix and plumbing.}
    \label{fig:app-qwen2b-technical-domain-delta}
\end{figure*}

To examine the effect of context on a single token's score, the examples in
Figure~\ref{fig:app-qwen2b-technical-domain-delta} hold the scored token fixed,
as in \Cref{fig:same-token-polysemy}. All four selected tokens rank near the
top of the vocabulary and have small reconstruction errors.\footnote{The
selected tokens rank \(17\)th or better, with relative reconstruction
error at most \(0.05\).}

\paragraph{Changing the score.}
\label{app:selected-readout-score-family-display-example}

The choice of score determines which token rows enter the decomposition
and how they are weighted (Appendix~\ref{app:score-decomposition}).

\begin{figure*}[!tbp]
    \centering
    \AppendixWideGraphic{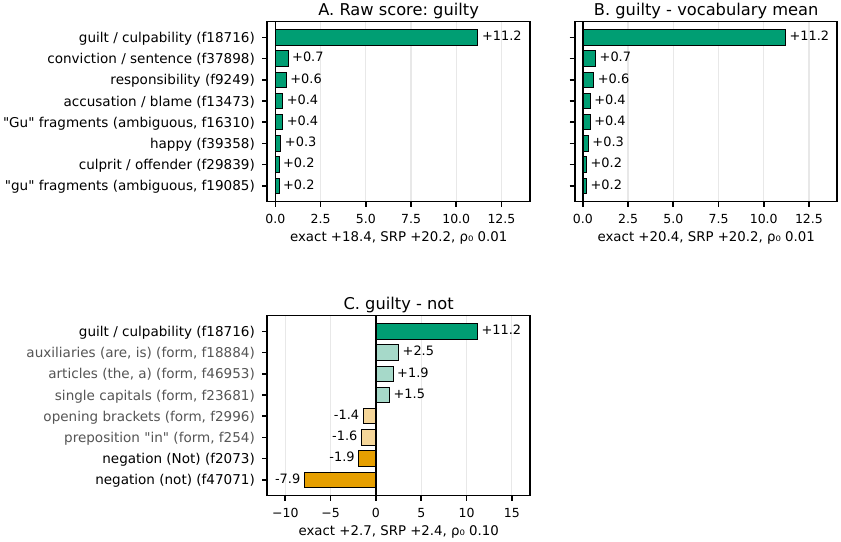}
    \caption{\textbf{Three scores on the same decoded state.}
    The prompt ends ``After reviewing the evidence, the jury found the
    defendant''. A feature associated with \texttt{guilty} and \texttt{guilt}
    supports all three scores. Features associated with \texttt{not}
    oppose the margin over \texttt{not}. Positive bars increase the score,
    and negative bars decrease it.}
    \label{fig:qwen-general-readout-scores-basic}
\end{figure*}

\begin{figure*}[!tbp]
    \centering
    \AppendixWideGraphic{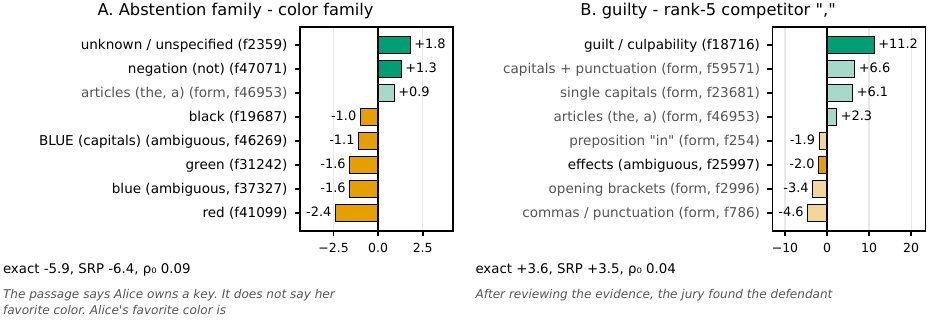}
    \caption{\textbf{A group contrast and a top-token margin.}
    On the left, color features help put the forced-answer group ahead of
    the abstention group. On the right, a feature associated with
    \texttt{guilty} and \texttt{guilt} gives the largest contribution
    in favor of the top token over its rank-5 competitor. Each label gives
    a short description of the top unembedding rows followed by the
    feature ID.}
    \label{fig:app-qwen-general-readout-scores-extra}
\end{figure*}

\subsection{Feature-Label Audit}
\label{app:main-case-study-feature-audit}

\paragraph{How labels are produced.}
A language model drafts short descriptions of each feature's
highest-scoring unembedding rows. In the preceding plots, labels for
individual features give a short description followed by the feature ID.
The labels mark token form or ambiguity where the audit identifies it.
One author checked whether each audited label described the top tokens
and whether the signed feature contribution supported the panel's
interpretation. Table~\ref{tab:app-qualitative-feature-label-audit}
summarizes the ratings. The code repository includes item-level annotations
for the two \texttt{bug} panels.

\paragraph{Blinded feature classification.}
As an additional check of how
consistently the features can be classified from their token rows alone,
we ran \texttt{gpt-5.6-sol} three times at high reasoning effort, with a
fresh context for each run and the same prompt. Each run saw only the top
20 rows per feature. The author labels, panel claims, and signed
contributions were hidden. The runs used the same item order, set by a
fixed random seed.

The three runs agreed unanimously on 93.4\% of the 91 distinct features.
Counting all 99 appearances in the audited displays, the
majority vote classified 76 as coherent, four as mixed or ambiguous, and
19 as token form. It agreed with the author's original three-way ratings
on 81.8\% of those appearances.
When the ratings differed, the majority vote usually classified the
feature as coherent.

\begin{table*}[!tbp]
\centering
\scriptsize
\setlength{\tabcolsep}{3pt}
\renewcommand{\arraystretch}{1.08}
\begin{tabularx}{\textwidth}{@{}lrrrrY@{}}
\toprule
Feature set & Labels & Readable & Token-form & On-target & Scope \\
\midrule
Main \texttt{bug} panels & 11 & 8 & 3 & 8 & Figure~\ref{fig:selected-readout-prism-examples}, detailed rows in Table~\ref{tab:app-main-case-study-feature-audit}. \\
\texttt{bark}/\texttt{bass} margins & 12 & 9 & 3 & 7 & Figure~\ref{fig:extended-four-margin-examples}. \\
Fixed-token context panels & 12 & 12 & 0 & 9 & Figure~\ref{fig:same-token-polysemy}. \\
Technical context stress & 32 & 31 & 1 & 24 & Figure~\ref{fig:app-qwen2b-technical-domain-delta}. \\
\texttt{verify}/\texttt{assume} DLA & 8 & 8 & 0 & 7 & Earlier eight-feature version of Figure~\ref{fig:qwen-prism-dla-feature-heatmap}. \\
General readout examples & 24 & 17 & 7 & 11 & Figures~\ref{fig:qwen-general-readout-scores-basic} and~\ref{fig:app-qwen-general-readout-scores-extra}. \\
\midrule
Total & 99 & 85 & 14 & 66 & Across the audited panels. \\
\bottomrule
\end{tabularx}
\caption{\textbf{Audit of labels in the Qwen figures.} In the two
\texttt{bug} panels, all readable labels support the panel's interpretation.
Readable combines the author's coherent and ambiguous categories. Token
forms include casing, leading spaces, subwords, and punctuation. On-target
labels have both a relevant token family and a contribution with the
appropriate sign.}
\label{tab:app-qualitative-feature-label-audit}
\end{table*}

The jury example shows a small positive term in the \texttt{guilty}
logit for the feature with top tokens \texttt{happy} and \texttt{happily}
(\Cref{fig:qwen-general-readout-scores-basic}). These tokens describe
positive emotion and give little evidence for the legal sense of
\texttt{guilty}, so the audit leaves the feature out of the on-target
count. The main text bases its panel interpretations on features that
pass both checks.

For Table~\ref{tab:app-main-case-study-feature-audit}, encoder scores on
centered, normalized unembedding rows determine the token rankings.

\begin{table*}[!tbp]
\centering
\scriptsize
\setlength{\tabcolsep}{3pt}
\renewcommand{\arraystretch}{1.08}
\begin{tabularx}{\textwidth}{@{}lY>{\RaggedRight\arraybackslash}p{0.24\textwidth}l@{}}
\toprule
Feature & Top associated rows, shortened & Interpretation & Confidence \\
\midrule
\texttt{f36} & \texttt{mosquito}, \texttt{mosquitoes}, \texttt{Mos}, \texttt{mos}, mosquito tokens in Chinese and Vietnamese, \texttt{insect}, \texttt{flea}, \texttt{malaria}, \texttt{bug}, \texttt{moth} & mosquito / insect / disease-vector bug sense & high \\
\texttt{f4095} & \texttt{bee}, \texttt{bees}, \texttt{Bee}, bee tokens in Chinese, \texttt{hive}, \texttt{butterfly}, \texttt{insect}, \texttt{ants}, \texttt{honey} & bees / insects & high \\
\texttt{f18302} & \texttt{Beetle}, \texttt{beetle}, \texttt{beet}, \texttt{Beet}, \texttt{Bug}, \texttt{bug}, \texttt{\_bug}, \texttt{BUG}, \texttt{buggy} & beetle / bug token family, mixed insect and software string forms & medium-high \\
\texttt{f58330} & \texttt{\_error}, \texttt{\_ERROR}, \texttt{error}, \texttt{-error}, \texttt{Error}, \texttt{.error}, \texttt{\_errors}, \texttt{/error}, \texttt{\_err} & programming/error-token feature & high \\
\texttt{f13081} & \texttt{Error}, \texttt{error}, \texttt{\_error}, \texttt{-error}, \texttt{ERROR}, \texttt{errors}, \texttt{.Error}, \texttt{\_Error}, Japanese error token & error / exception surface forms & high \\
\texttt{f59571} & \texttt{M}, \texttt{J}, \texttt{G}, \texttt{K}, \texttt{N}, \texttt{P}, \texttt{I}, \texttt{L}, \texttt{A}, \texttt{B}, \(\ldots\), newline-like token & single leading-space capitals / low-level formatting-like feature & low \\
\texttt{f21804} & \texttt{defect}, \texttt{defects}, defect tokens in Chinese and Russian, \texttt{flaw}, \texttt{defe}, \texttt{defective}, \texttt{flaws}, \texttt{bug}, \texttt{fault} & defect / flaw / software-bug sense & high \\
\texttt{f43419} & \texttt{Crash}, \texttt{crash}, \texttt{crashes}, \texttt{crashed}, \texttt{crashing}, \texttt{Bug}, \texttt{bug}, \texttt{bugs}, \texttt{panic} & crash / bug / panic software failure & high \\
\texttt{f5680} & \texttt{debug}, \texttt{Debug}, \texttt{.debug}, \texttt{\_debug}, \texttt{DEBUG}, \texttt{debugging}, \texttt{debugger}, \texttt{dbg}, Chinese debug token & debug / debugging code tokens & high \\
\texttt{f34693} & \texttt{p}, \texttt{j}, \texttt{d}, \texttt{b}, \texttt{t}, \texttt{g}, \texttt{m}, \texttt{l}, \texttt{h}, \texttt{s}, \texttt{c}, \texttt{n} & single lowercase leading-space letters & low \\
\texttt{f23681} & \texttt{M}, \texttt{P}, \texttt{S}, \texttt{K}, \texttt{G}, \texttt{A}, \texttt{D}, \texttt{B}, \texttt{N}, \texttt{J}, \texttt{I}, \texttt{Y}, \texttt{H}, \texttt{O}, \texttt{C}, \texttt{F} & single uppercase letters / alphabetic token family & low \\
\bottomrule
\end{tabularx}
\caption{\textbf{Unembedding rows behind the \texttt{bug} feature labels.}
The table lists each distinct feature in Figure~\ref{fig:selected-readout-prism-examples}.
The top rows include both sense-related tokens and spelling variants.
Leading spaces are omitted, and non-Latin synonyms are summarized.}
\label{tab:app-main-case-study-feature-audit}
\end{table*}

\subsection{Word-Sense Prediction}\label{app:sense-labelled-evaluation}

Recall features contribute more to the \texttt{memory} logit in a
sentence about human memory, and a RAM feature contributes more in a
sentence about computers (Appendix~\ref{app:additional-margin-case-studies}).
Both scores use the same readout row. Their contributions differ because
the two decoded states align differently with the features in that row.

The held-out evaluation asks whether these patterns recur often enough
for small feature groups to distinguish senses in new contexts.

\paragraph{Data and scoring.}
The evaluation uses the fixed train and test split of CoarseWSD-20
\citep{loureiro2021coarsewsd} and the \(32\times\), \(k=256\) dictionary for
each model. We use one cloze template and score the known ambiguous word
at the final prompt position regardless of its rank. Both Qwen models
encode every target word as a single token. R1-Llama-8B is evaluated
only on the words it encodes as single tokens.\footnote{The dataset
contains 33{,}566 contexts, of which 16{,}785 are eligible for R1-Llama-8B.
The target row has relative reconstruction error below 0.5 in every Qwen
context and 97.6\% of R1-Llama-8B contexts. Median target ranks are 174 on
Qwen3.5-2B, 10 on Qwen3.5-9B, and 1{,}131 on R1-Llama-8B.}

\paragraph{Group size and baseline comparisons.}
For the groups in \Cref{tab:sense-alignment-summary}, we rescale each
feature's contributions on both splits using its mean and standard
deviation on the training contexts for that word. At eight features per
sense, every word on Qwen3.5-2B and all but one on each other model beat
the 95th percentile of the shuffled-label control.\footnote{Every word
beats its majority baseline at each tested group size. The majority
baseline is 0.410 for both Qwen models and 0.429 for R1-Llama-8B. Across
group sizes one, two, four, and eight, the number of words above the
shuffled-label threshold changes by at most one. Ordinary accuracy gives
the same ordering.}

\subsection{Stability Across Independent Training Runs}
\label{app:basis-stability}

Do token groups for a contrast recur when we retrain the dictionary?
At each of two width and sparsity settings, we train three Qwen3.5-2B
dictionaries from scratch. Preprocessing stays fixed, while the seeds for
initialization, data order, and sampling vary.\footnote{Validation top-1
agreement ranges from 0.805 to 0.848.}

\paragraph{Token groups for the same contrast.}
\label{app:basis-stability-explanations}

For a contrast between \(A\) and \(B\), we pool the top vocabulary tokens
from the features that most favor \(A\) over \(B\) and do the same for
the features that favor \(B\). This gives two token groups per dictionary.
Between runs, we compare groups for the same side of a contrast, for
opposite sides, and for unrelated contrasts. We measure Jaccard overlap,
the fraction of distinct tokens across the two groups that appear in both.

\begin{figure}[!tbp]
\centering
\includegraphics[width=\columnwidth]{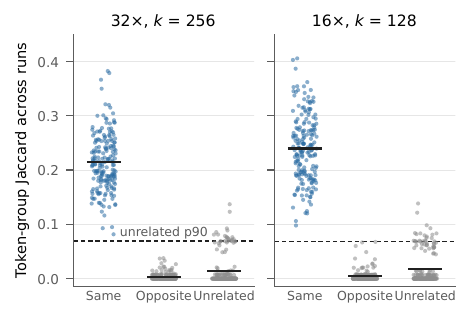}
\caption{\textbf{Token-group overlap across training runs.} Each dot
compares two token groups from a pair of Qwen3.5-2B runs at the same width.
Bars mark means, and the dashed line marks the 90th percentile for unrelated
contrasts. Mean overlap for the same side is 0.214 at \(32\times\) and 0.240
at \(16\times\).}
\label{fig:app-cross-seed-overlap}
\end{figure}

For all 63 contrasts and every pair of runs, the mean of the two same-side
overlaps exceeds the 90th percentile for unrelated contrasts
(\Cref{fig:app-cross-seed-overlap}). Groups for opposite sides share few
tokens. Decoder directions vary across runs, with lower similarity between
matched features at the larger width (\Cref{tab:app-cross-seed-stability}),
consistent with earlier results on TopK dictionaries
\citep{paulo2025stability}.

\begin{table}[!tbp]
\centering
\footnotesize
\setlength{\tabcolsep}{4pt}
\renewcommand{\arraystretch}{1.08}
\begin{tabularx}{\linewidth}{@{}>{\RaggedRight\arraybackslash}Xcc@{}}
\toprule
Metric & \(32\times\)/\(k{=}256\) & \(16\times\)/\(k{=}128\) \\
\midrule
Matched-projection correlation & 0.53--0.54 & 0.63--0.66 \\
Matched decoder cosine \mbox{(used features)} & 0.31--0.34 & 0.52--0.55 \\
\bottomrule
\end{tabularx}
\caption{\textbf{Agreement of decoder directions across training runs.}
Ranges span the three pairs of Qwen3.5-2B runs at each width. Projection
correlation uses held-out states projected onto the features used in the
contrasts and their closest decoder matches.}
\label{tab:app-cross-seed-stability}
\end{table}

\paragraph{Token groups for individual features.}
\label{app:basis-stability-groups}

Does a feature's token group reappear together in another run, or are
its tokens spread across several features? For each pair of runs, we
sample 100 features used in the contrasts. Each group consists of the
feature's 12 highest-scoring tokens from the centered unembedding rows.

We search all \(D\) groups in the other dictionary. Best-single Jaccard
is the largest overlap with any one group. Recall@3 is the fraction of
the original tokens recovered by up to three groups, adding the group
that covers the most remaining tokens at each step. As a chance baseline,
we repeat the search for 500 random groups with similar token frequencies.

In \Cref{tab:app-feature-group-matching}, most groups have a single
match above the chance threshold. Combining up to three groups recovers
most of the original tokens. Some decoder directions match almost
exactly, and matches tend to be closer in the narrower dictionary,
consistent with \citet{paulo2025stability}.

\begin{table}[!tbp]
\centering
\footnotesize
\setlength{\tabcolsep}{4pt}
\renewcommand{\arraystretch}{1.08}
\begin{tabularx}{\linewidth}{@{}>{\RaggedRight\arraybackslash}p{0.52\columnwidth}ZZ@{}}
\toprule
Metric (range over pairs) & \(32\times\) & \(16\times\) \\
\midrule
Best-single Jaccard, median & 0.32--0.42 & 0.37--0.43 \\
Groups above null p99 & 89--90\% & 87--91\% \\
Strong equivalence (\(J\ge0.5\)) & 36--41\% & 40--47\% \\
Recall@3, median & 0.83--0.90 & 0.86--0.88 \\
Recall@3 above null p99 & 91--93\% & 92\% \\
Matched decoder cosine, \mbox{med / p90} & 0.20--0.26 / 0.80--0.89 & 0.40--0.42 / 0.90--0.94 \\
\bottomrule
\end{tabularx}
\caption{\textbf{Matching feature groups across training runs.} Ranges span
the three pairs of Qwen3.5-2B runs at each width. For the null groups, the
median best-single Jaccard is 0.059 and its 99th percentile is
\(\approx\) 0.11. Null recall@3 is 0.25.}
\label{tab:app-feature-group-matching}
\end{table}

Some features with low token overlap still have a similar decoder
direction in the other dictionary.\footnote{Of the 32 and 33 groups below
the null out of 300 at the two widths, 5 and 10 have a decoder counterpart at cosine
\(\ge 0.5\), respectively.} On informal review, most of the remaining
features resembled those with mixed or ambiguous labels in the audit.
The threshold-based counts use fixed cutoffs and are descriptive.

\label{app:basis-stability-loo}
\label{app:basis-stability-cross-recipe}
The released dictionary used earlier preprocessing and recovered 36\%
of the tokens that appeared on the same side of a contrast in at least
two of the three runs under the newer recipe.

\subsection{Reading the Logit Lens Across Layers}
\label{app:all-layer-srp-lens}

Layerwise lenses show how token predictions change through a model
\citep{nostalgebraist2020,belrose2023tuned}, and Logit Prisms break scores
into contributions from model components \citep{nguyen2024logitprisms}.
At each layer and position, the logit lens reads the full hidden state
to give the top token and its probability. For selected cells, SRP uses
the same decoded state to show which readout features support or oppose
a score for that token. The displays sum contributions from features
with similar top unembedding rows. On Qwen3.5-2B (\(D=65536\), \(k=256\)),
the score is the top token's vocabulary-mean contrast. On Ministral-3-8B,
it is the top token's logit with the contribution of the learned offset
\(\mu\) subtracted.

\paragraph{From a category to a name.}
Panel A of \Cref{fig:app-all-layer-srp-lens-ministral-paris} follows
Ministral's predictions after ``The Eiffel Tower is located in the city
of''. The lens first predicts \texttt{city} and later \texttt{Paris}.
Cases 1--3 in panel B show the change in feature contributions. A group
for cities in general supports \texttt{city}. Groups for France and major
cities already support \texttt{Paris} when it first becomes the top token,
and a group for the spelling \textit{par} is prominent at a later layer.
Across 19 city rows, the major-city feature ranks in the top three for
16 cities, and a feature for the matching country ranks in the top four
for 17. Features for other countries fall outside the top four.\footnote{We
identify countries by names and adjectives among a feature's 20 top
unembedding rows, and exclude features that name more
than two of the 13 countries. The cities are Paris, Lyon, Berlin, Munich,
Hamburg, Madrid, Barcelona, Rome, Milan, Tokyo, Osaka, Moscow, Vienna,
London, Cairo, Lisbon, Chicago, Toronto, and Sydney.}

\paragraph{Calendar names and their positions.}
Panel B of \Cref{fig:app-all-layer-srp-lens-ministral-days} shows a group
for weekday names supporting \texttt{Wednesday}, \texttt{Thursday}, and
\texttt{Friday}, alongside groups for three, four, and five, respectively.
The pattern also
appears in the unembedding rows. In 12 of 19 weekday and month rows, a
feature for the matching calendar position ranks above features for
other numbers (\Cref{tab:app-ministral-calendar-numbers}).\footnote{A
feature is specific to a number in this check if it ranks in the top 16
for that number word alone among the words for one through twelve.}

\begin{figure*}[!tbp]
    \centering
    \includegraphics[width=\textwidth,height=0.84\textheight,keepaspectratio]{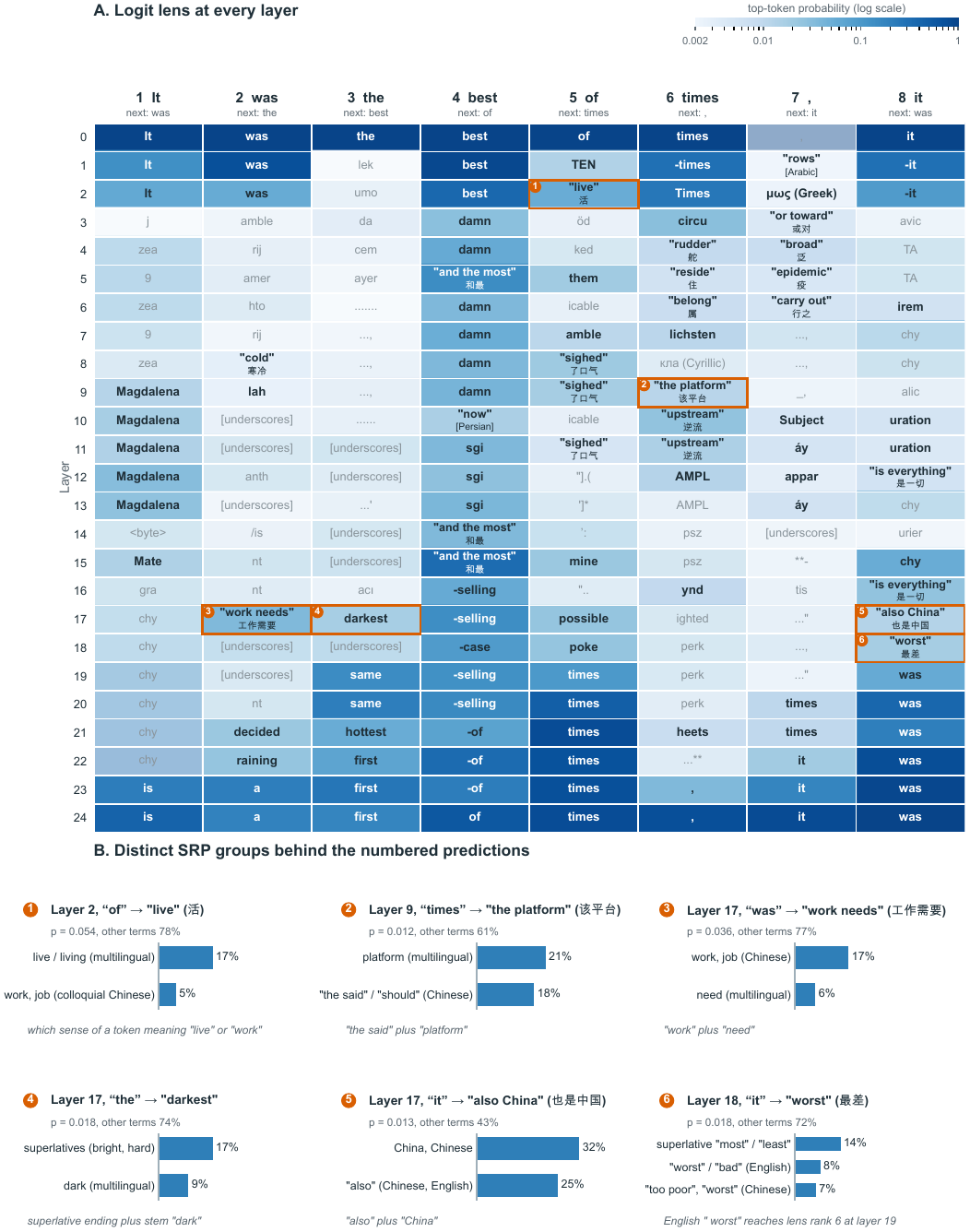}
    \caption{\textbf{Logit-lens predictions and readout features across layers.}
    \textbf{(A)} The Qwen3.5-2B logit lens gives the top token at each
    layer and position. Shading shows its probability on a log scale.
    Grey marks subword or formatting predictions that differ from the
    correct next token, shown below the input. English glosses appear
    above non-Latin tokens. \textbf{(B)} For each numbered cell, SRP merges
    similar features and shows groups that carry at least 4\% of the
    positive contribution. In these examples, the groups can
    distinguish senses or account for separate parts of a token.}
    \label{fig:app-all-layer-srp-lens}
\end{figure*}

\begin{figure*}[!tbp]
    \centering
    \includegraphics[width=\textwidth,height=0.84\textheight,keepaspectratio]{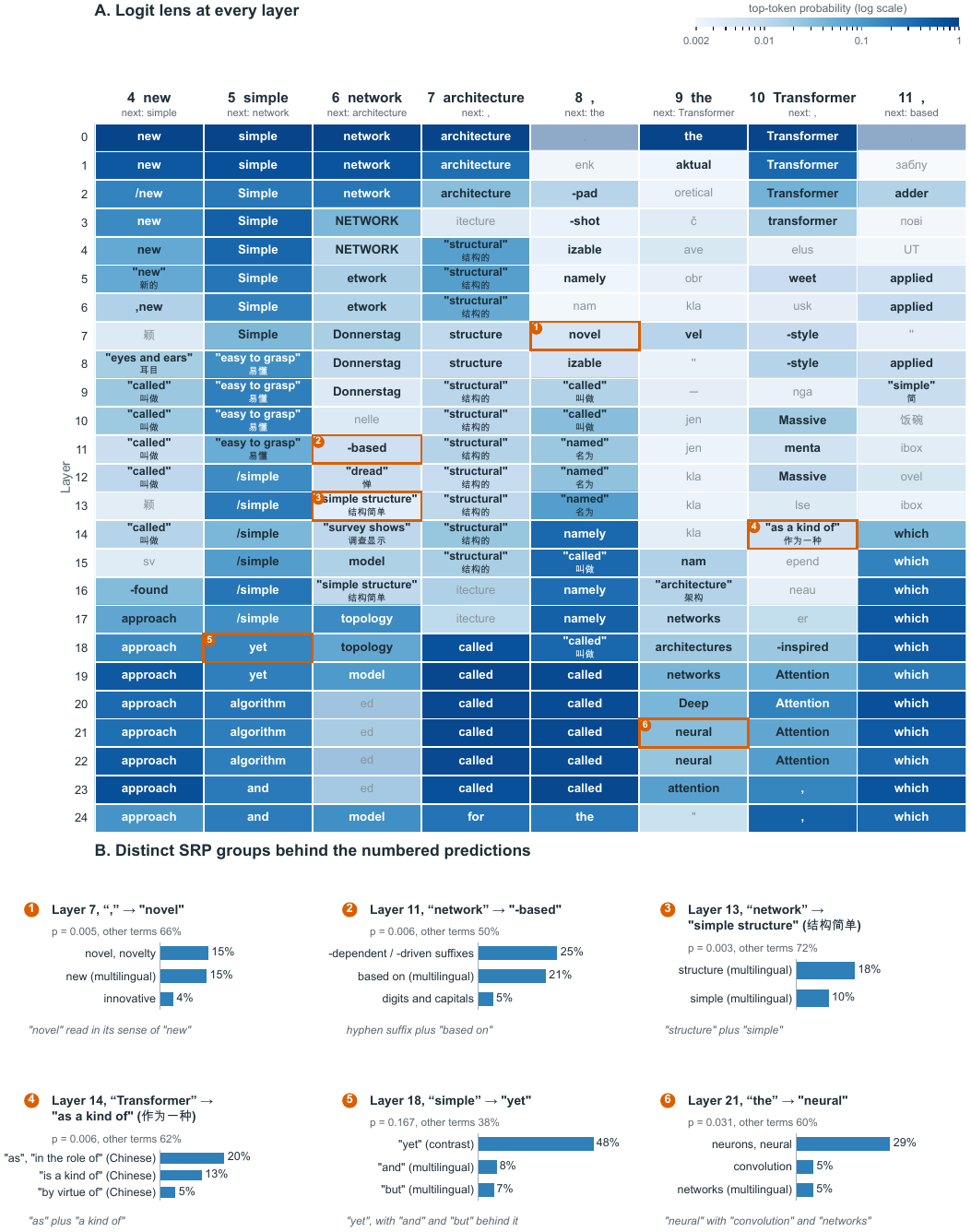}
    \caption{\textbf{Readout features for the Transformer prompt.}
    The panels follow \Cref{fig:app-all-layer-srp-lens} for positions 4 to 11
    of ``We propose a new simple network architecture, the Transformer,
    based solely on attention mechanisms''. In one numbered cell,
    separate groups support the two parts of the token for ``simple structure''.}
    \label{fig:app-all-layer-srp-lens-transformer}
\end{figure*}

\begin{figure*}[!tbp]
    \centering
    \includegraphics[width=\textwidth,height=0.84\textheight,keepaspectratio]{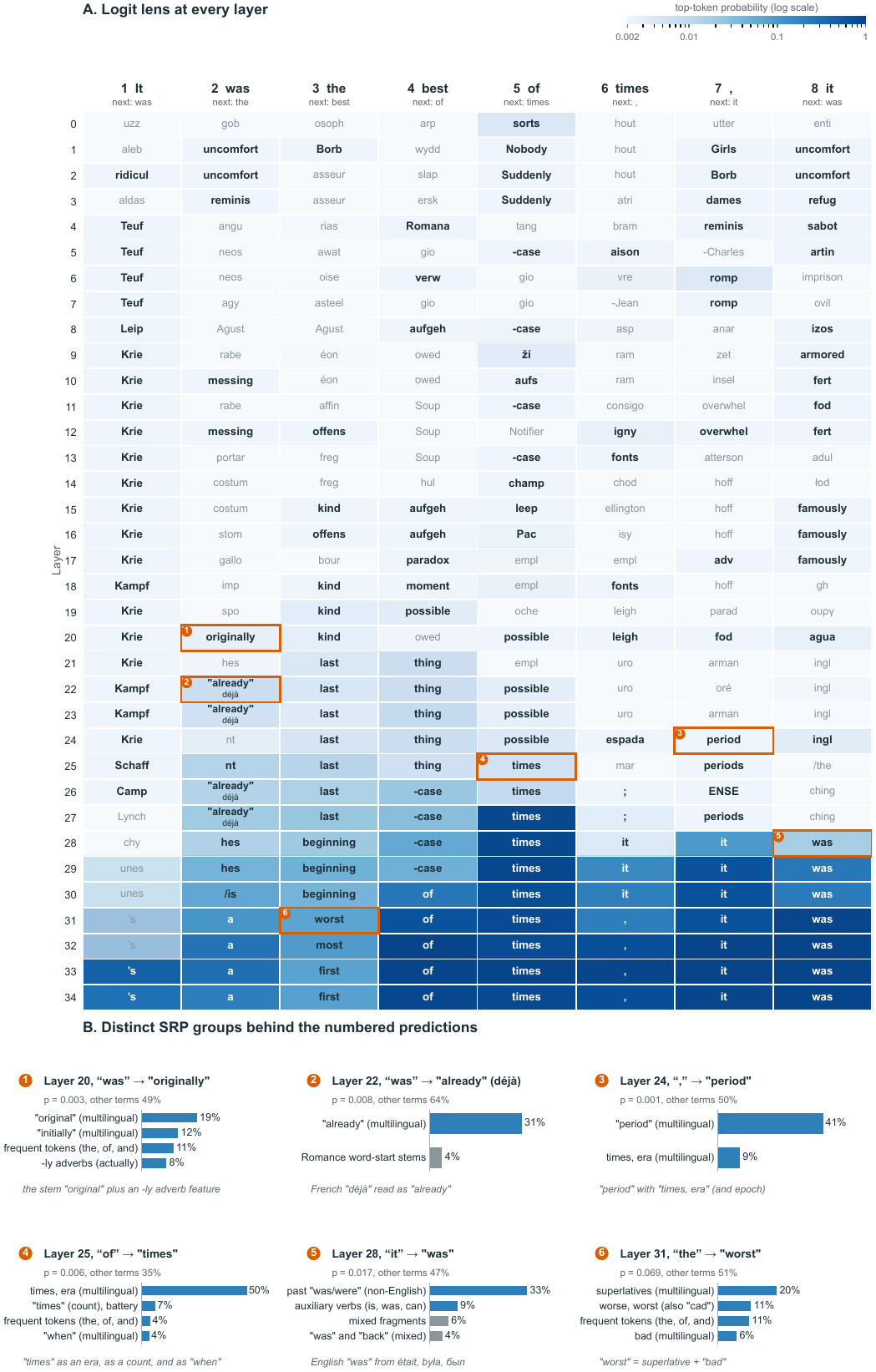}
    \caption{\textbf{Readout features for the Dickens prompt on Ministral.}
    The panels follow \Cref{fig:app-all-layer-srp-lens} for Ministral-3-8B
    with \(D=131072\), \(k=256\). Positions start after the beginning-of-sequence
    token. English glosses also appear for \textit{d\'ej\`a} and \textit{einf}.
    Panel B keeps the frequent-token feature separate and shows up to
    four groups per cell down to 3\%. For \texttt{worst}, the groups
    distinguish superlative features from features for bad.}
    \label{fig:app-all-layer-srp-lens-ministral}
\end{figure*}

\begin{figure*}[!tbp]
    \centering
    \includegraphics[width=\textwidth,height=0.84\textheight,keepaspectratio]{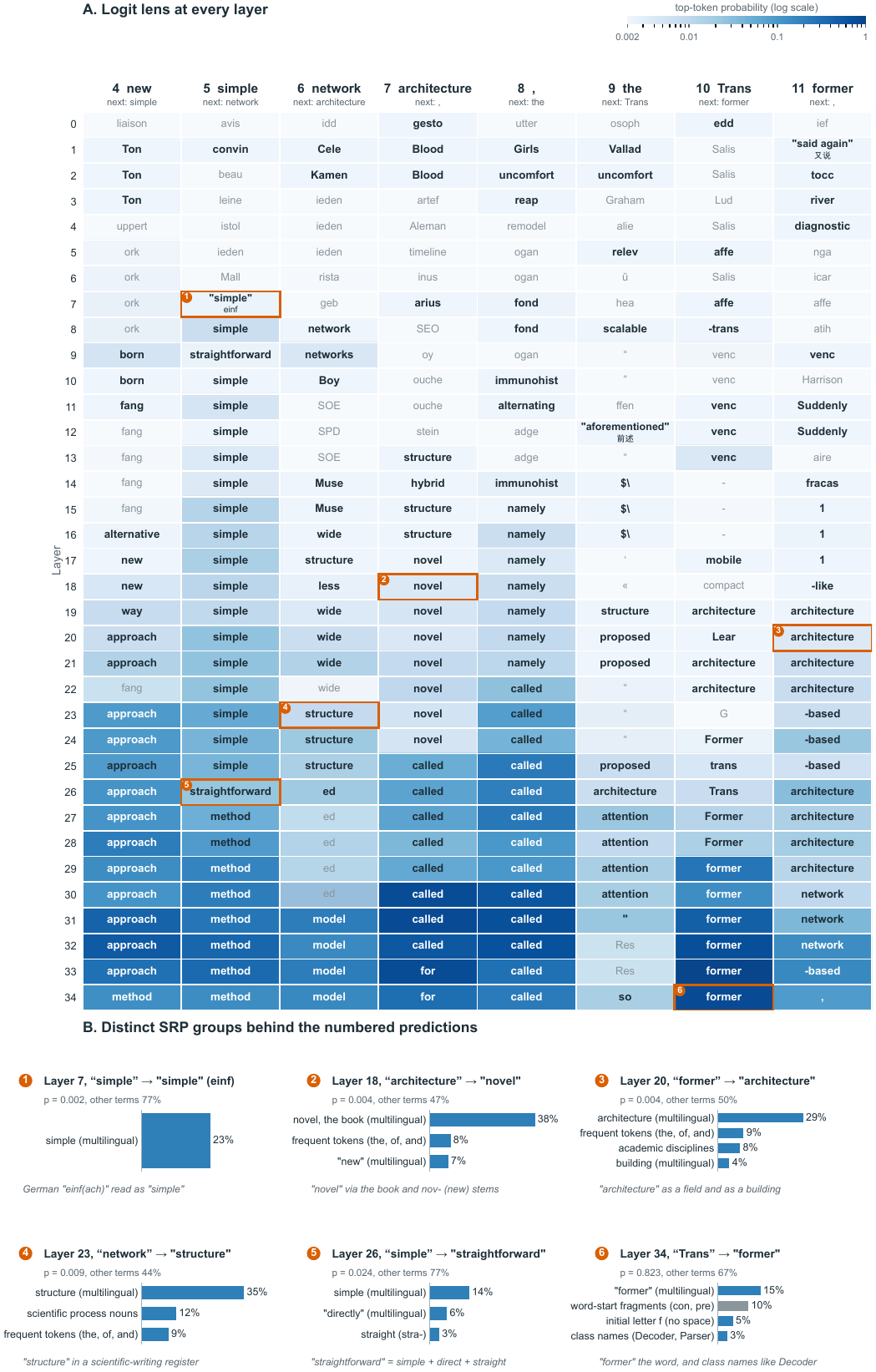}
    \caption{\textbf{Readout features for the Transformer prompt on Ministral.}
    The panels follow \Cref{fig:app-all-layer-srp-lens-ministral} for positions
    4 to 11 of ``We propose a new simple network architecture, the Transformer,
    based solely on attention mechanisms''. The features for \texttt{architecture}
    include groups for fields of study and buildings.}
    \label{fig:app-all-layer-srp-lens-ministral-transformer}
\end{figure*}

\begin{figure*}[!tbp]
    \centering
    \includegraphics[width=\textwidth,height=0.84\textheight,keepaspectratio]{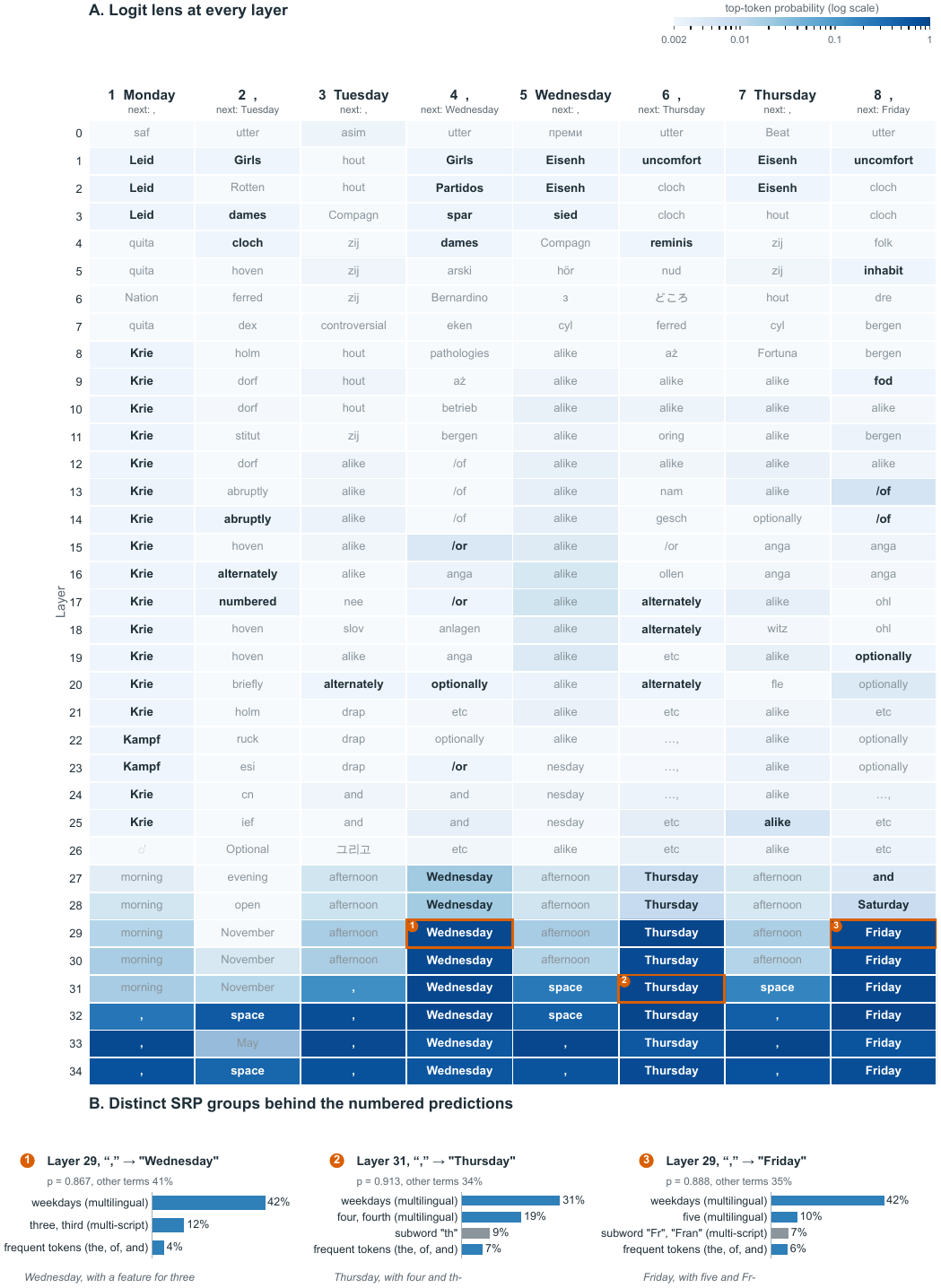}
    \caption{\textbf{Readout features for the weekdays prompt on Ministral.}
    The panels follow \Cref{fig:app-all-layer-srp-lens-ministral} for
    ``Monday, Tuesday, Wednesday, Thursday, Friday''. Each predicted weekday
    draws on a group for weekday names and a group for its position in the
    week.}
    \label{fig:app-all-layer-srp-lens-ministral-days}
\end{figure*}

\begin{figure*}[!tbp]
    \centering
    \includegraphics[width=\textwidth,height=0.84\textheight,keepaspectratio]{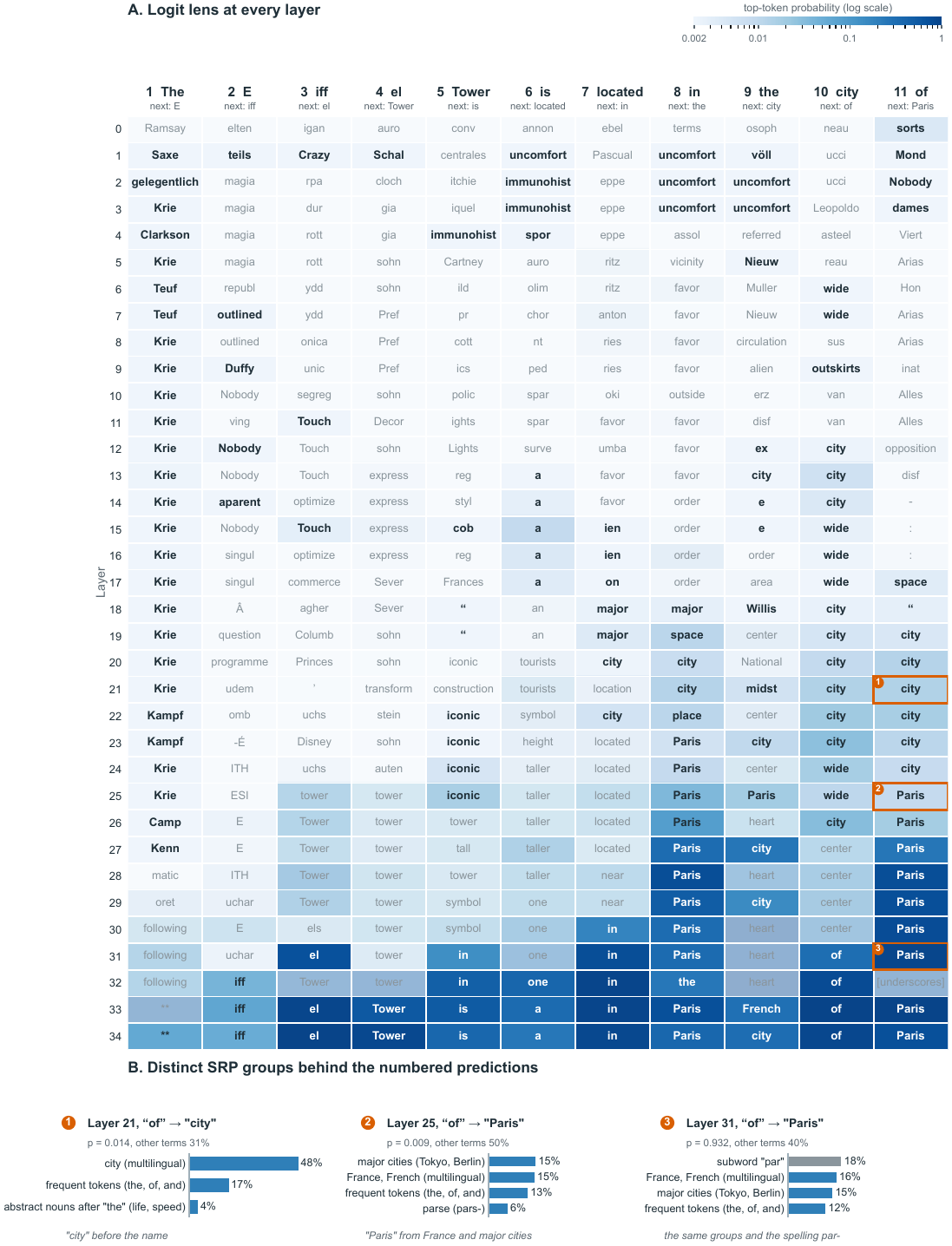}
    \caption{\textbf{Readout features for the Eiffel Tower prompt on Ministral.}
    The panels follow \Cref{fig:app-all-layer-srp-lens-ministral} for
    ``The Eiffel Tower is located in the city of Paris''. Separate groups for
    France, major cities, and the spelling \textit{par} support
    \texttt{Paris}.}
    \label{fig:app-all-layer-srp-lens-ministral-paris}
\end{figure*}

Case 6 in both Dickens plots
(\Cref{fig:app-all-layer-srp-lens,fig:app-all-layer-srp-lens-ministral})
shows that Qwen's Chinese token for ``worst'' and Ministral's English
\texttt{worst} both draw on separate groups for the superlative and for
``bad''. In case 2 of \Cref{fig:app-all-layer-srp-lens-transformer},
Qwen's \texttt{-based} draws on one group for hyphenated endings and
another for ``based on''. Cases 1 and 5 of
\Cref{fig:app-all-layer-srp-lens-ministral-transformer} show that
Ministral's German fragment \textit{einf} and its later prediction
\texttt{straightforward} both draw on features for simplicity at the same
position. \Cref{tab:app-chinese-tokens} gives glosses for the Chinese
tokens in the Qwen displays.

\begin{table}[!tbp]
\centering
\scriptsize
\setlength{\tabcolsep}{3pt}
\renewcommand{\arraystretch}{1.1}
\begin{tabularx}{\linewidth}{@{}l>{\RaggedRight\arraybackslash}p{0.25\linewidth}Y@{}}
\toprule
Token & English & Parts and meaning \\
\midrule
\begin{CJK*}{UTF8}{gbsn}活\end{CJK*} & live & Means ``alive'' or ``to live'', and in \begin{CJK*}{UTF8}{gbsn}活儿\end{CJK*} also a job or a piece of work. \\
\begin{CJK*}{UTF8}{gbsn}该平台\end{CJK*} & the platform & \begin{CJK*}{UTF8}{gbsn}该\end{CJK*} is a formal ``the said'' or ``this'' and on its own also means ``should''. \begin{CJK*}{UTF8}{gbsn}平台\end{CJK*} means platform. \\
\begin{CJK*}{UTF8}{gbsn}工作需要\end{CJK*} & work needs & \begin{CJK*}{UTF8}{gbsn}工作\end{CJK*} means work or job, and \begin{CJK*}{UTF8}{gbsn}需要\end{CJK*} means need or require, so the token reads ``the job requires''. \\
\begin{CJK*}{UTF8}{gbsn}也是中国\end{CJK*} & also China & One token joins three words, \begin{CJK*}{UTF8}{gbsn}也\end{CJK*} (also), \begin{CJK*}{UTF8}{gbsn}是\end{CJK*} (is), and \begin{CJK*}{UTF8}{gbsn}中国\end{CJK*} (China), and reads ``is also China''. \\
\begin{CJK*}{UTF8}{gbsn}最差\end{CJK*} & worst & \begin{CJK*}{UTF8}{gbsn}最\end{CJK*} marks the superlative and \begin{CJK*}{UTF8}{gbsn}差\end{CJK*} means poor or bad. \\
\begin{CJK*}{UTF8}{gbsn}结构简单\end{CJK*} & simple structure & \begin{CJK*}{UTF8}{gbsn}结构\end{CJK*} means structure and \begin{CJK*}{UTF8}{gbsn}简单\end{CJK*} means simple. The adjective follows the noun, so the token reads ``simple in structure''. \\
\begin{CJK*}{UTF8}{gbsn}作为一种\end{CJK*} & as a kind of & \begin{CJK*}{UTF8}{gbsn}作为\end{CJK*} means ``as'' or ``in the role of''. \begin{CJK*}{UTF8}{gbsn}一种\end{CJK*} is ``one kind'', a number with the classifier that Chinese places between a number and a noun. \\
\bottomrule
\end{tabularx}
\caption{\textbf{Chinese tokens in the layer-by-layer displays.} The table
gives the English gloss used in
\Cref{fig:app-all-layer-srp-lens,fig:app-all-layer-srp-lens-transformer}
and shows how each token splits into parts.}
\label{tab:app-chinese-tokens}
\end{table}

\begin{table}[!tbp]
\centering
\scriptsize
\setlength{\tabcolsep}{3pt}
\begin{tabular}{@{}lrrr@{\hspace{10pt}}lrrr@{}}
\toprule
 & & \multicolumn{2}{c}{Rank} & & & \multicolumn{2}{c}{Rank} \\
\cmidrule(lr){3-4}\cmidrule(lr){7-8}
Weekday & \(n\) & \(n\) & Other & Month & \(n\) & \(n\) & Other \\
\midrule
Monday & 1 & -- & -- & January & 1 & -- & 160 \\
Tuesday & 2 & 15 & 100 & February & 2 & -- & -- \\
Wednesday & 3 & 3 & 154 & March & 3 & 18 & 124 \\
Thursday & 4 & 3 & -- & April & 4 & 7 & 134 \\
Friday & 5 & 7 & -- & May & 5 & 68 & 126 \\
Saturday & 6 & 18 & -- & June & 6 & 2 & 63 \\
Sunday & 7 & -- & 210 & July & 7 & 1 & 54 \\
 & & & & August & 8 & 21 & 171 \\
 & & & & September & 9 & 63 & 164 \\
 & & & & October & 10 & -- & -- \\
 & & & & November & 11 & -- & -- \\
 & & & & December & 12 & -- & -- \\
\bottomrule
\end{tabular}
\caption{\textbf{Number features in Ministral's weekday and month rows.}
For each row, the table gives the rank of the feature for its calendar
position \(n\), counted from Monday and January, among the 256 features the
row uses. It also gives the best rank of a feature for any other number
from one to twelve. A dash means the row uses no such feature.}
\label{tab:app-ministral-calendar-numbers}
\end{table}

\section{Directional Ablation}
\label{app:readout-interventions}
\label{app:causal-validation}
\label{subsec:causal-validation}

\paragraph{Predicted and measured drops.}
Directional ablation tests how closely a feature's contribution predicts
the effect of removing its direction from the decoded state.
For a contrast direction \(q_\alpha\) and decoded state \(\rstate\),
SRP assigns feature \(i\) the contribution \(c_i=\beta_i(\alpha)\,p_i(\rstate)\).
Ablation removes the component of the state along the unit-norm direction
\(d_i\), giving \(\rstate' = \rstate - (\rstate^\top d_i)\,d_i\).
We express the score change \(s_\alpha(\rstate')-s_\alpha(\rstate)\) as a
signed drop, so positive values correspond to a decrease in the score.
\[
    D_i := s_\alpha(\rstate)-s_\alpha(\rstate')
    =p_i(\rstate)\,d_i^\top q_\alpha.
\]
The shared offset cancels for a contrast. Expanding the readout direction
then gives
\[
\begin{aligned}
    D_i &= c_i + p_i(\rstate)\Big(
        \textstyle\sum_{j\ne i}\beta_j(\alpha)\,d_i^\top d_j \\
        &\hspace{6.2em}{}+d_i^\top r_\alpha\Big).
\end{aligned}
\]
Removing one direction can also change the contributions of features
whose directions overlap with it, as well as the score from the readout
residual. Both effects can reinforce or offset the selected feature's
contribution, so the score change can differ from \(-c_i\).

\paragraph{Regression and controls.}
The test uses the \(k=256\) dictionaries for the six linear readouts.
For each contrast, we ablate the ten features with the largest absolute
contributions and ten features sampled at random from the remaining
directions (\Cref{tab:causal-validation-summary}). The random features can
have zero coefficients, and sampling does not match contribution magnitudes.
For each model, we regress the measured drop on the feature contribution
through the origin for contrasts that meet the coverage criterion
(\S\ref{sec:method-reading-local-accounts}). The \(r^2\) measures how closely
the contributions track the drops, and a slope of one means they have the
same scale. Including all pairs gives similar fits. A synthetic check with
zero residual gives a slope close to one and zero predicted contributions
for the controls.\footnote{Including all pairs changes \(r^2\) by less than 0.02
for each model. The synthetic check gives \(r^2=1.000\) and a slope of 1.04.}

\begin{table}[!tbp]
\centering
\footnotesize
\setlength{\tabcolsep}{3.2pt}
\renewcommand{\arraystretch}{1.06}
\begin{tabular*}{\linewidth}{@{\extracolsep{\fill}}lcccc@{}}
\toprule
Model & \(r^2\) [95\% CI] & Slope & Random & \(n\) \\
\midrule
Q-0.8B & 0.886 [.869, .903] & 1.661 & 0.010 & 1610 \\
Q-2B   & 0.834 [.802, .862] & 1.338 & 0.007 & 1820 \\
Q-9B   & 0.829 [.813, .845] & 1.686 & 0.058 & 1860 \\
Min-8B & 0.926 [.913, .937] & 0.957 & 0.009 & 2050 \\
R1-Q-7B & 0.918 [.899, .933] & 1.082 & 0.000 & 1500 \\
R1-L-8B & 0.934 [.920, .947] & 1.126 & 0.034 & 1210 \\
\bottomrule
\end{tabular*}
\caption{\textbf{Changes in the margin after directional ablation.}
Selected contributions track the measured drops more closely than the
random controls. The regression fits the signed drop against the feature
contribution through the origin at a fixed decoded state. Each model
starts with \(\sim\)260 contrasts, and \(n\) counts ablations on contrasts
that meet the coverage criterion. Random gives \(r^2\) for 10 features
sampled outside each selected set. Brackets give bootstrap 95\% intervals.}
\label{tab:causal-validation-summary}
\end{table}

\paragraph{Results.}
The selected contributions track the measured drops more closely than
the controls on every model. Slopes are close to one on Ministral and
the two R1-distilled models. On Qwen3.5 the slopes are above one, so the
contributions tend to underestimate the size of the drops
(\Cref{tab:causal-validation-summary}).

\paragraph{Separating overlap and residual effects.}
% Run records: ../../work/camera_ready/experiments/ablation_slope_20261118/{qwen0p8b,qwen2b}/summary.json
On Qwen3.5-0.8B and Qwen3.5-2B, a follow-up asks why the measured drops
tend to be larger than the feature contributions. Both runs center the
readout rows using the mean across the full vocabulary and use contrasts
from the A/B and case-candidate banks. Fitting the overlap and residual
terms separately against \(c_i\) through the origin shows how far each
term shifts the slope. Together these shifts account for the difference
between the fitted slope and one.

\begin{table}[!tbp]
\centering
\footnotesize
\setlength{\tabcolsep}{2pt}
\renewcommand{\arraystretch}{1.08}
\begin{tabular*}{\linewidth}{@{\extracolsep{\fill}}lccc@{}}
\toprule
Model & Slope & Overlap shift & Residual shift \\
\midrule
Q-0.8B & 1.662 & \makecell{\(+0.678\)\\\([+0.629,+0.728]\)} & \makecell{\(-0.016\)\\\([-0.024,-0.008]\)} \\
Q-2B & 1.338 & \makecell{\(+0.345\)\\\([+0.312,+0.381]\)} & \makecell{\(-0.007\)\\\([-0.011,-0.003]\)} \\
\bottomrule
\end{tabular*}
\caption{\textbf{Decoder overlap and the residual in directional ablation.}
Overlap accounts for the slopes above one on both models. The residual
shifts each slope slightly downward. Each run uses a \(32\times\),
\(k=256\) dictionary and 264 contrasts. The runs include 1,590 and 1,820
covered feature--contrast pairs, respectively. Brackets give 95\% intervals
from 2,000 bootstrap resamples of base-case clusters.}
\label{tab:ablation-slope-decomposition}
\end{table}

On both models, decoder overlap accounts for the larger fitted drops,
and the residual offsets a small part of that increase
(\Cref{tab:ablation-slope-decomposition}).
The same pattern holds when all contrasts are included. In a spot check,
the drop from the equation agrees with scoring both states directly using
the LM head.\footnote{With all contrasts, overlap shifts are
\(+0.649\) and \(+0.325\), and residual
shifts are \(-0.017\) and \(-0.005\), respectively. On 100 ablations per
model, the analytic drop agrees with direct scoring within
\(5.0\times10^{-7}\) logits.}

\paragraph{Scope.}
The ablation changes the decoded state only for scoring with the original
LM head. The measured drop describes how that score responds to removing
a direction at the selected state.

\FloatBarrier
\section{Feature-Resolved Direct Logit Attribution}
\label{app:feature-resolved-dla}

\begin{figure*}[!tbp]
    \centering
    \includegraphics[width=\textwidth]{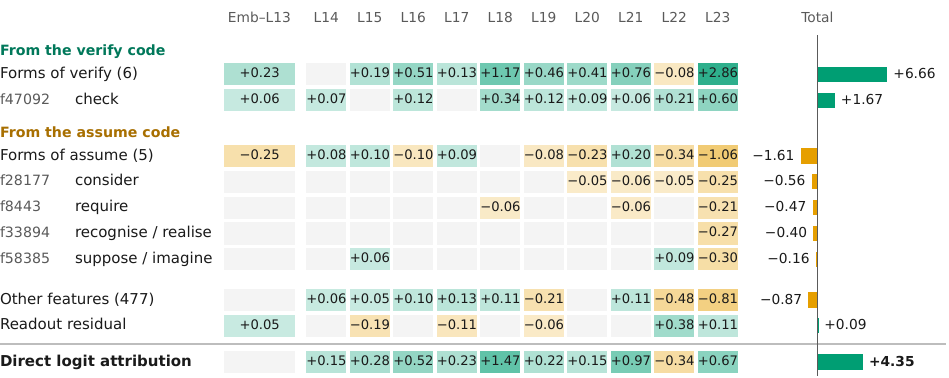}
    \caption{\textbf{Direct logit attribution by readout feature.}
    Rows show feature contributions and the readout residual for the final
    Qwen3.5-2B \texttt{verify} minus \texttt{assume} margin. Columns show
    residual-stream components, with the embedding and layers 0--13 pooled
    on the left and the total across components on the right. The bottom
    row gives ordinary direct logit attribution. Green favors
    \texttt{verify}, orange favors \texttt{assume}, and component cells below
    \(0.05\) logits in magnitude are blank. We group features by the token
    with the larger SAE coefficient and pool the features associated with
    forms of that token in one row per group. Of the remaining features,
    those with the largest summed absolute contributions get their own
    rows, and the rest share the \emph{Other features} row.}
    \label{fig:qwen-prism-dla-feature-heatmap}
\end{figure*}

We use the readout basis from \S\ref{sec:results} to trace how individual
features contribute to the final margin across layers. Ordinary direct
logit attribution (DLA) estimates each residual-stream component's
contribution to that margin
\citep{elhage2021framework,wang2022ioi,nguyen2024logitprisms}.
We take the embedding and each layer's update to the residual stream as
components \(j\). Applying the final normalization with its RMS denominator
held fixed gives the component vectors \(\tilde h_{L,j}\), whose sum
approximates the decoded state \(\rstateL\) at the final layer \(L\),
\(\rstateL\approx\sum_j \tilde h_{L,j}\). We apply the score decomposition
from \S\ref{sec:method-local-score} to each component. For a pairwise margin,
the shared offset vanishes because the token weights \(\alpha\) sum to zero.
The remaining terms are the contribution \(M_{j,i}(\alpha)\) of feature \(i\)
and the readout residual term \(R_j(\alpha)\),
\[
    \begin{aligned}
    M_{j,i}(\alpha)&=\beta_i(\alpha)\,\tilde h_{L,j}^\top d_i,\\
    R_j(\alpha)&=\tilde h_{L,j}^\top r_\alpha .
    \end{aligned}
\]
The feature direction \(d_i\) has coefficient \(\beta_i(\alpha)\) in the
selected readout direction, and \(r_\alpha\) is the SAE reconstruction
residual of that direction.

In \Cref{fig:qwen-prism-dla-feature-heatmap}, Qwen3.5-2B reads ``The source
has not been checked yet. The appropriate next action is to''.
We use the \(32\times\), \(k=256\) dictionary. For this margin, \(\alpha\)
assigns \(+1\) to \texttt{verify}, \(-1\) to \texttt{assume}, and zero
elsewhere. In each column, the feature contributions and the readout
residual sum to the DLA value before the display rounds values and hides
small entries.

Ordinary DLA assigns \(+0.7\) logits to the final layer. Verification and
checking features account for \(+3.5\) logits at that layer, most of which
is offset by features for \texttt{assume} and related verbs and by other
features.
The total DLA score exceeds the exact margin of \(+4.31\) logits by \(0.04\) logits,
and the readout SAE residual accounts for \(+0.09\) logits of the DLA total.

\section{Lexical Editing on Held-Out Tokens}
\label{app:lexical-editing-matched-kl}

We select SRP features using five profanity tokens and test the edit on
other tokens from the same category. Each test pairs one of these tokens
with an alternative next token.

\paragraph{Transfer to held-out tokens.}
The SRP edit subtracts a scaled sum of the selected feature directions
from the state passed to the LM head. The edit makes the alternative
score higher in all 23 cases where Qwen3.5-2B initially favors the
profanity token. Directly lowering the scores of the five tokens used
to select features changes the preferred token in zero of those cases.

\paragraph{Comparing edit directions.}
We compare SRP with three simple baselines built from the same five
tokens. One averages their unembedding rows, and two use PCA. For each
pair, we measure how much the edit lowers the probability of choosing
the profanity token over its alternative. We compare edits near four
KL targets to match how much they change the full next-token
distribution.

\begin{samepage}
On Qwen3.5-2B, SRP lowers the test token's probability more than each
baseline at all four KL targets. Its gain over the mean-row edit ranges
from 3.2 to 8.7 percentage points (\Cref{fig:lexical-editing-matched-kl-effects}).
All 12 paired confidence intervals on Qwen3.5-2B exclude zero.
The mean-row edit gives the larger drop at all four targets on
Qwen3.5-0.8B and Qwen3.5-9B, and at two targets on R1-Distill-Qwen-7B.\par
\end{samepage}

\medskip
\noindent\begin{minipage}{\columnwidth}
    \centering
    \includegraphics[width=\linewidth]{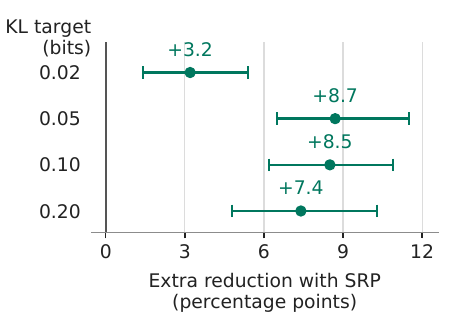}
    \captionof{figure}{\textbf{Extra reduction with SRP on Qwen3.5-2B.}
    Near the 0.05-bit KL target, SRP lowers the test token's probability
    by 8.7 percentage points more than the mean-row baseline. Bars show
    paired 95\% confidence intervals.}
    \label{fig:lexical-editing-matched-kl-effects}
\end{minipage}

\end{document}